\documentclass[runningheads]{llncs}

\usepackage{eccv}

\usepackage{multirow}
\usepackage{eccvabbrv}

\usepackage{graphicx}
\usepackage{booktabs}

\usepackage[accsupp]{axessibility}  
\usepackage{threeparttable}  

\usepackage[pagebackref,breaklinks,colorlinks,citecolor=eccvblue]{hyperref}

\usepackage{orcidlink}

\begin{document}

\title{Camera-LiDAR Cross-modality Gait Recognition} 

\titlerunning{CL-Gait}

\author{Wenxuan Guo$^*$\inst{1}\orcidlink{0009-0007-4823-1587} \and 
Yingping Liang$^*$\inst{2}\orcidlink{0000-0001-5385-0015} \and
Zhiyu Pan\inst{1}\orcidlink{0009-0000-6721-4482} \and
Ziheng Xi\inst{1}\orcidlink{0009-0008-7007-8803} \and
Jianjiang Feng$^\dag$\inst{1}\orcidlink{0000-0003-4971-6707} \and
Jie Zhou\inst{1}\orcidlink{0000-0001-7701-234X}}

\authorrunning{W. Guo, Y. Liang et al.}

\institute{Department of Automation, Tsinghua University, China \and
Beijing Institute of Technology\\
\email{\{gwx22, pzy20, xizh21\}@mails.tsinghua.edu.cn}
\email{  liangyingping@bit.edu.cn}\\
\email{\{jfeng, jzhou\}@tsinghua.edu.cn}}

\maketitle

\begin{abstract}
Gait recognition is a crucial biometric identification technique. Camera-based gait recognition has been widely applied in both research and industrial fields. LiDAR-based gait recognition has also begun to evolve most recently, due to the provision of 3D structural information. However, in certain applications, cameras fail to recognize persons, such as in low-light environments and long-distance recognition scenarios, where LiDARs work well. On the other hand, the deployment cost and complexity of LiDAR systems limit its wider application. Therefore, it is essential to consider cross-modality gait recognition between cameras and LiDARs for a broader range of applications. In this work, we propose the first cross-modality gait recognition framework between \textbf{C}amera and \textbf{L}iDAR, namely \textbf{CL-Gait}. It employs a two-stream network for feature embedding of both modalities. This poses a challenging recognition task due to the inherent matching between 3D and 2D data, exhibiting significant modality discrepancy. To align the feature spaces of the two modalities, i.e., camera silhouettes and LiDAR points, we propose a contrastive pre-training strategy to mitigate modality discrepancy. To make up for the absence of paired camera-LiDAR data for pre-training, we also introduce a strategy for generating data on a large scale. This strategy utilizes monocular depth estimated from single RGB images and virtual cameras to generate pseudo point clouds for contrastive pre-training. Extensive experiments show that the cross-modality gait recognition is very challenging but still contains potential and feasibility with our proposed model and pre-training strategy. To the best of our knowledge, this is the first work to address cross-modality gait recognition.
\end{abstract}    
\section{Introduction}
\label{sec:intro}

\begin{figure*}[t]
	\begin{center}
		\begin{subfigure}{0.48\linewidth}
			\begin{center}
				\includegraphics[width=\linewidth]{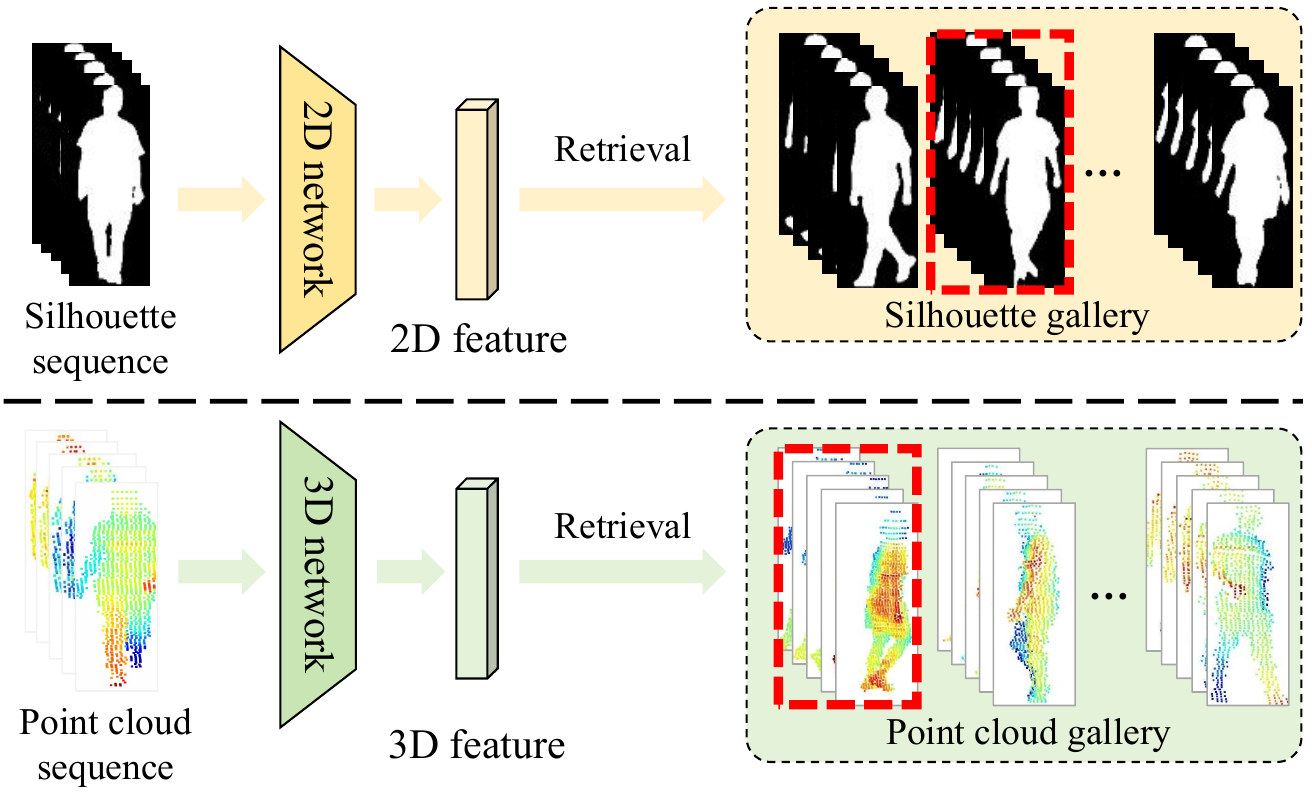}
				\caption{Camera-based gait recognition (top), and LiDAR-based gait recognition (bottom).}\label{fig:intro1}
			\end{center}
		\end{subfigure}
		\quad
		\begin{subfigure}{0.48\linewidth}
			\begin{center}
				\includegraphics[width=\linewidth]{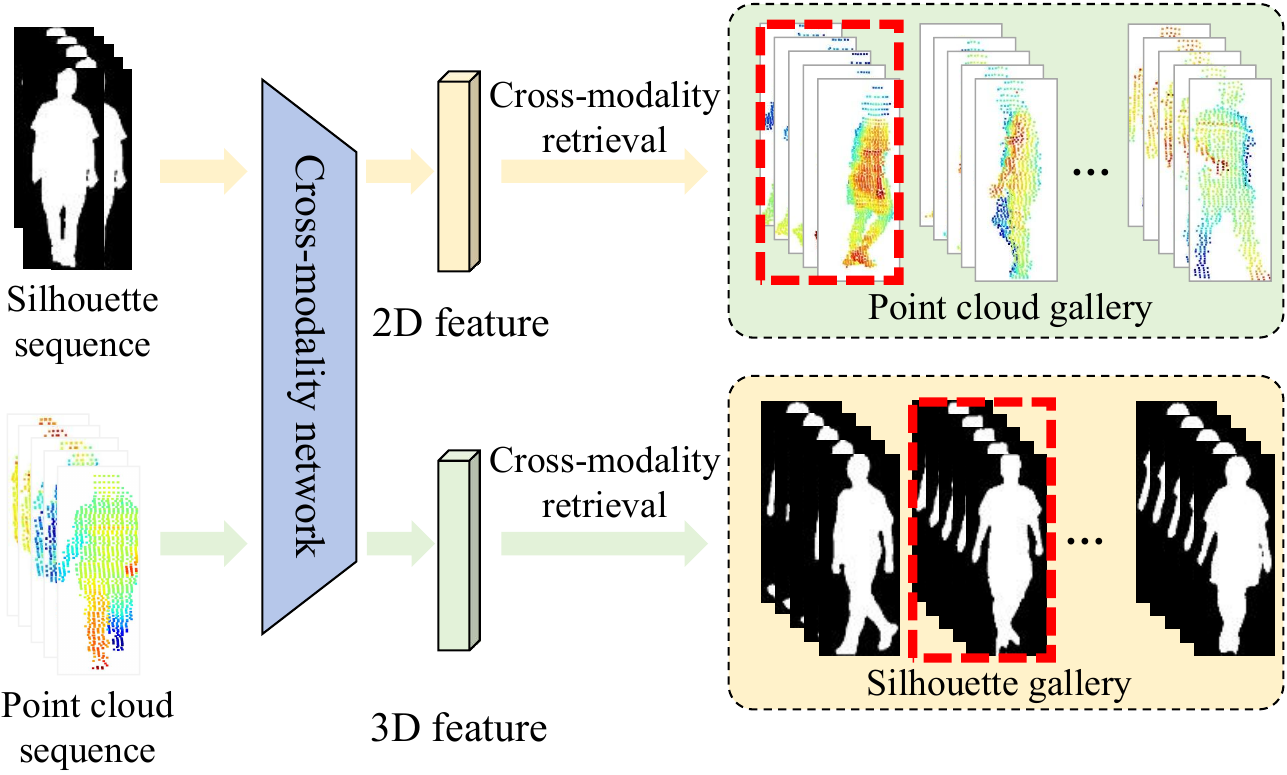}
				\caption{Our proposed cross-modality gait recognition between camera and LiDAR.}
				\label{fig:intro2}
			\end{center}
		\end{subfigure}	
	\end{center}
	\vspace{-.2cm}
	\caption{Overview of single-modality and cross-modality gait recognition. Single-modality gait recognition takes data of one modality as input and searches within a gallery of the same data type. In contrast, cross-modality gait recognition processes two modalities and identify individuals within a gallery of a different modality.}\label{fig:com}
\end{figure*}

Gait recognition is a crucial long-range identification technology without physical contact, which has great advantages in privacy protection and cross-clothing recognition \cite{chao2019gaitset}. It identifies individuals based on behavioral characteristics extracted from walking sequences recorded by sensors. Gait recognition has been extensively applied in fields such as user identification, sports science, and public security. Most current gait recognition research employs cameras as sensors \cite{sepas2022deep}. With the advancement of computer vision technology, camera-based gait recognition continues to achieve new progress \cite{lin2022gaitgl,liang2022gaitedge,lin2021gaitmask,liao2020model,teepe2021gaitgraph,loper2023smpl}.

While utilizing cameras for gait recognition is cost-effective, it faces limitations in certain applications, such as low-light environments, long-distance recognition, and scenarios requiring precise 3D perception. In low-light conditions, the available information in RGB images significantly diminishes \cite{guo2023lidar}, making it challenging to accurately detect pedestrians and segment their silhouettes. Furthermore, due to the decline in image clarity, camera-based systems struggle to distinguish the fine-grained features from a distance \cite{fan2023opengait}. Additionally, accurately capturing the 3D movement of individuals, crucial for comprehensive gait analysis, is beyond the capability of traditional camera systems \cite{shen2023lidargait}.

Recently, to address the challenges under low light environment faced by RGB cameras, some researches \cite{ling2020class,ye2020cross,park2021learning,lu2020cross} have focused on RGB-Infrared cross-modality person re-identification (ReID), achieving commendable performance. However, infrared cameras struggle to capture pedestrians wearing insulating or thick clothing. Besides, they are constrained by limited detection distance and resolution, as well as susceptibility to lighting conditions and weather. It also faces the risk of pedestrian privacy and safety, compared with gait recognition. These limitations highlight the need for exploring alternative or complementary technologies that can offer more robust solutions for person identification tasks across a variety of environmental conditions.

In recent years, LiDAR technology has been increasingly utilized \cite{wang2019pseudo} due to its impressive gains on 3D information. Beyond its mature applications in autonomous driving, LiDAR has also been applied in individual identification areas, such as gait recognition \cite{shen2023lidargait} and person re-identification \cite{guo2023lidar} most recently. Through point cloud data, LiDAR can provide precise 3D perception, aiding in the acquisition of an individual's 3D geometric structure. From this, intrinsic characteristics of a person, including height, body shape, and gait, can be extracted. Moreover, LiDARs possess long-distance perception capabilities, unaffected by lighting conditions and complex backgrounds. And it offers enhanced privacy protection than camera \cite{guo2023lidar}.  The robustness to environmental variables and the ability to capture detailed biometric data make it an invaluable tool for advanced surveillance and security systems. However, compared to camera systems, LiDAR's deployment complexity and cost are somehow higher, making it unsuitable for large-scale deployment in general scenarios.

Considering the respective advantages of cameras and LiDARs,  switching sensors in different scenarios to explore cross-modality gait recognition presents a worthwhile task for investigation. Cross-modality gait recognition offers an option with a broader application scope. Cameras can be utilized in general scenes with normal lighting, while LiDAR can be applied in specific scenarios, offering a complementary tool that enhances overall recognition capabilities.

In this work, we present the first study on camera-LiDAR cross-modality gait recognition as shown in Fig.~\ref{fig:com}. This task poses significant challenges due to the need for matching data between 3D and 2D modalities. Based on our observations, there are substantial differences between point clouds and images. To be specific, for gait recognition, point clouds primarily focus on the 3D positioning of body parts, whereas images capture precise contour information of individuals. To address these challenges and explore the feasibility of cross-modality gait recognition, we propose a novel cross-modality framework, namely CL-Gait. 

To extract gait features from both modalities, CL-Gait basically employs a two-stream network. Considering modality differences and data processing consistency, we first project and upsample the point clouds to obtain depth images with the same resolution as silhouettes, and use the  projected depth images to train the network. Furthermore, to mitigate modality discrepancy, CL-Gait utilizes a contrastive silhouette-point pre-training approach (CSPP) to align the feature spaces of the two modalities. Pre-training requires aligned camera-LiDAR data under the same view, which is hard to obtain. To this end, we propose a new strategy of generating cross-modality gait data for pre-training. This strategy utilizes advances in monocular depth estimation and is able to generate large-scale dataset for pre-training using only single RGB images. 

Extensive experiments have revealed the following \textbf{three} insights: \textbf{1)} CL-Gait achieves an average rank-1 accuracy of $54.21\%$. The results show at least $22.90\%$ improvement over the baseline models and demonstrate the significant potential of cross-modality gait recognition. \textbf{2)} Utilizing large-scale depth images generated from point clouds as input is proved superior to using point clouds directly. \textbf{3)} The strategy of contrastive pre-training on large-scale generated paired data mitigates the modality differences, contributing to performance improvement.


To summarize, our main contributions are as follows:
\begin{itemize}
	\item[$\bullet$] To the best of our knowledge, this is the first work on cross-modality gait recognition. Extensive experiments show the potential of utilizing camera and LiDAR for gait recognition in challenging cross-modality scenarios.
	\item[$\bullet$] We analyse several network structures and conduct comparable experiments on their effectiveness for cross-modality gait recognition. Based on the two-stream network with better performance, our proposed CL-Gait is capable of matching camera silhouettes and LiDAR points.
	\item[$\bullet$] We propose a contrastive pre-training method to align the feature spaces of the two modalities. To make up for the absence of paired camera-LiDAR data, we further introduce a large-scale data generation strategy.
\end{itemize}

\section{Related Work}
\label{sec:related}
\noindent\textbf{Gait Recognition.}
Based on the sensor used, exiting gait recognition methods can be categorized into camera-based and LiDAR-based methods. 

Camera-based gait recognition has been extensively studied over the past decade. The majority of camera-based methods focus on extracting appearance features directly from images or videos~\cite{chao2019gaitset,fan2020gaitpart}, adapting well to resolution reduction and achieving impressive performance~\cite{fan2023opengait}. These methods utilize silhouettes~\cite{lin2022gaitgl,liang2022gaitedge,lin2021gaitmask} or other gait templates~\cite{bobick2001recognition,wang2010chrono} for spatial feature extraction and temporal modeling. Additionally, some researchers have explored using estimated underlying structure of human body~\cite{li2020end,liao2020model,teepe2021gaitgraph}, such as 2D/3D pose and the SMPL model~\cite{loper2023smpl}, as inputs. While theoretically robust against factors like carrying and clothing, these models often struggle at low resolutions, limiting their practicality in some real-world scenarios~\cite{fan2023opengait}.

LiDAR-based gait recognition is an emerging field that utilizes precise 3D representations, point clouds, capturing complex motion patterns and the 3D structure of individual gaits. This approach is less susceptible to variations in lighting, clothing, and background, offering promising avenues for accurate gait recognition in diverse conditions. LidarGait~\cite{shen2023lidargait} stands as a pioneering work in this field, introducing a multimodal gait recognition dataset named SUSTECH1K. This dataset is designed for evaluating the performance of gait recognition based on various sensors and has demonstrated the potential and practicality of LiDAR-based gait recognition. Additionally, some researchers have applied LiDAR to the task of person re-identification~\cite{guo2023lidar}, achieving impressive results.

Camera-based and LiDAR-based methods have been proven practical, each with its unique advantages and disadvantages. Therefore, providing a compromise solution by addressing cross-modality gait recognition represents a valuable research direction, which has not been studied currently.

\noindent\textbf{Contrastive Pre-training.}
Contrastive pre-training is primarily applied in the field of vision-language models, aimed at aligning the feature spaces of visual and linguistic models. In recent years, inspired by the success of self-supervised learning within intra-modal tasks, researchers have begun to explore pre-training objectives for tasks involving multiple modalities, such as vision and language~\cite{yang2022vision}. The pioneering work by CLIP~\cite{radford2021learning} performs cross-modality contrastive pre-training on hundreds of millions of image-text pairs. CLIP~\cite{radford2021learning} can generate a task-agnostic model that achieves surprisingly effective results. Subsequently, ALIGN~\cite{jia2021scaling} expands upon CLIP by utilizing a noisy dataset that covers more than a billion image-text pairs, further extending the capabilities of cross-modality pre-training. Additionally, MDETR~\cite{kamath2021mdetr} trains an end-to-end model on existing multimodal datasets, achieving explicit alignment between phrases in texts and objects in images. 
Contrastive pre-training plays a crucial role in cross-modality tasks. Therefore, we introduce contrastive pre-training into camera-LiDAR cross-modality gait recognition.

\noindent\textbf{RGB-IR Cross-modality Person ReID.}
Visible-infrared person ReID addresses the challenge of matching persons across different modalities, specifically between RGB and infrared cameras~\cite{wu2017rgb}. This task is particularly significant for scenarios with poor illumination, notably at night. With two primary datasets, SYSU-MM01~\cite{wu2017rgb} and RegDB~\cite{nguyen2017person}, RGB-IR cross-modality person ReID has been extensively studied, most of which focusing on metric learning, feature learning, and adversarial learning approaches. Metric learning~\cite{feng2019learning,hao2019hsme,ling2020class,ye2020cross} and feature learning~\cite{park2021learning,ye2019modality,ye2020dynamic} methods aim to extract and align multimodal features into a common space for effective comparison, employing strategies like angle-based measurement and Euclidean constraints to bridge the modality gap. Notably, advancements in this area include the development of novel loss functions~\cite{ling2020class} that enhance discrimination between modalities while preserving identity information. Concurrently, Generative Adversarial Networks (GANs)~\cite{choi2020hi,dai2018cross,lu2020cross} have been instrumental in cross-modality ReID, enabling the transformation of images between RGB and IR while maintaining identity information, and fostering feature learning through adversarial and disentanglement strategies. Despite their potential, GAN-based methods face challenges such as high computational demands and the occasional production of low-quality images that may negatively impact ReID performance. 

While RGB-IR person ReID mainly relies on appearance for distinction, which is a short-term feature and may raise privacy concerns. In contrast, cross-modality gait recognition focuses on shape and gait of individuals, which are time-invariant features, and offers better protection of privacy.
\section{Method}
\label{sec:method}
\begin{figure}
	\centering
	\includegraphics[width=0.95\linewidth]{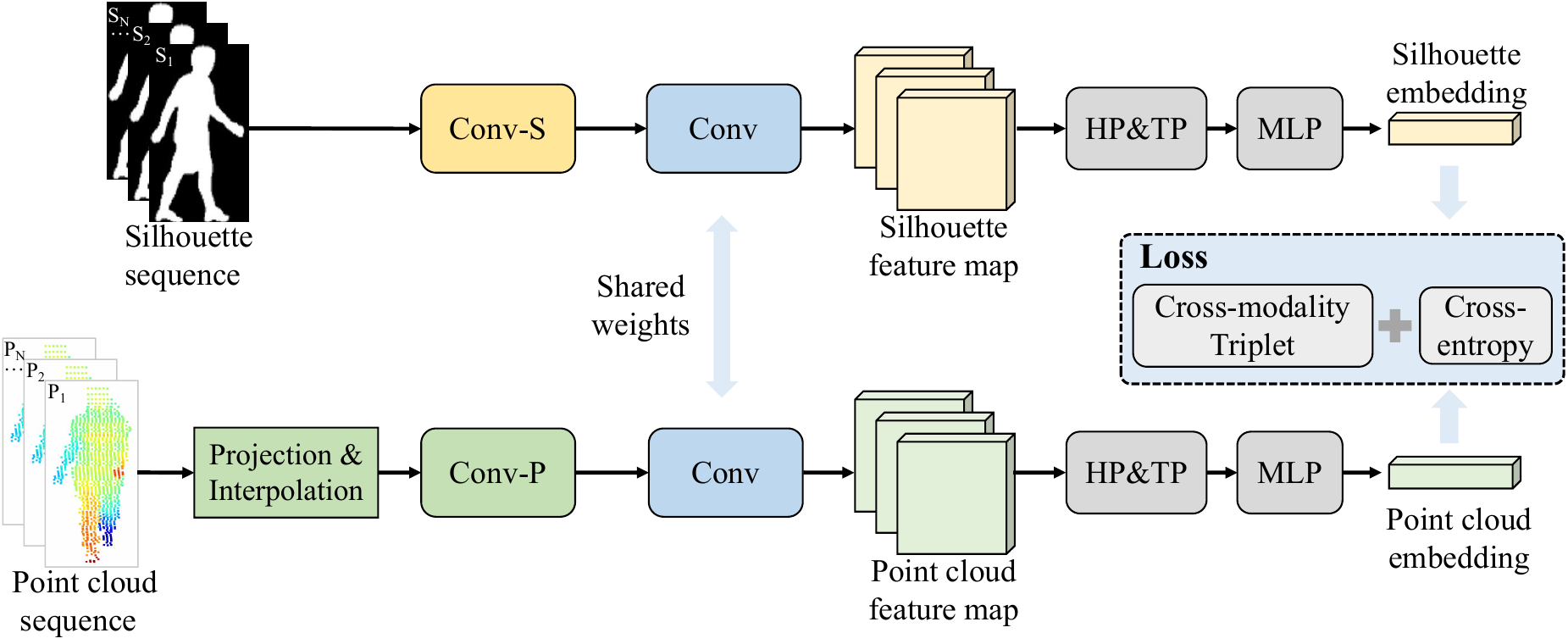}
	\caption{The cross-modality network of our proposed CL-Gait. It employs a two-stream architecture that encodes sequences from two modalities into a consistent feature space. HP stands for horizontal pooling, and TP represents temporal pooling.}
	\label{fig:method}
\end{figure}

\begin{figure}[t]
	\centering
	\includegraphics[width=0.95\linewidth]{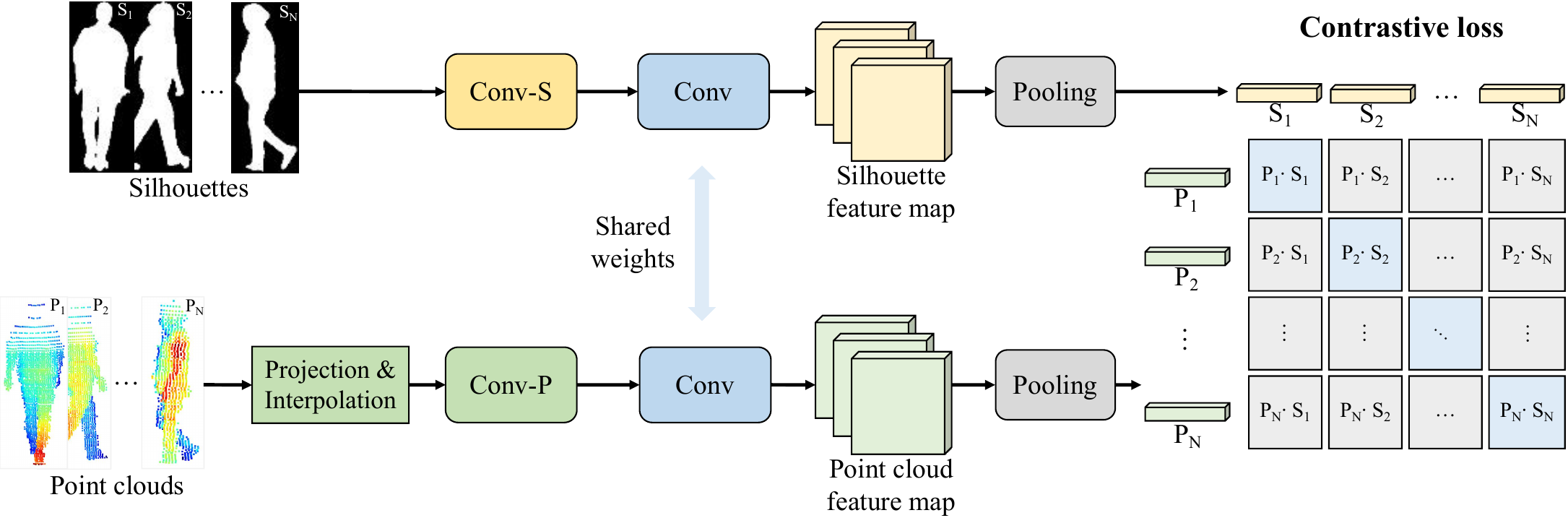}
	\caption{The contrastive silhouette-point pre-training (CSPP) approach of CL-Gait. Paired silhouettes and point clouds are taken as inputs, and the backbone of the cross-modality network is pre-trained with contrastive learning loss to align the feature spaces of the two modalities. The pre-training process does not require identity labels of the samples for supervision.}
	\label{fig:method_pre}
\end{figure}

In this work, we propose CL-Gait for cross-modality gait recognition between camera and LiDAR. CL-Gait employs a two-stream network for cross-modality feature embedding. The network utilizes modality-specific modules in the shallow layers and shared modules in the deeper layers, as illustrated in Fig. \ref{fig:method}. Besides, CL-Gait adopts a contrastive learning strategy to align the feature spaces of the two modalities, mitigating modality differences, as shown in Fig. \ref{fig:method_pre}. The backbone obtained from pre-training is used to initialize the cross-modality embedding network. To facilitate pre-training on large-scale data, we also propose a method for generating cross-modality gait data, as shown in Fig. \ref{fig:cspp}.

\subsection{Cross-modality Network of CL-Gait}
\noindent\textbf{Two-Stream Backbone.} 
The CL-Gait framework processes two distinct data modalities: 2D silhouette sequences and 3D point cloud sequences, optimizing cross-modality feature extracting. Inspired by \cite{shen2023lidargait}, we first project and interpolate point clouds into depth images. For each modality, we apply modality-specific convolution layers (Conv-S for silhouettes and Conv-P for depth images generated from point clouds) before projecting the features into a shared embedding space, as shown in Fig. \ref{fig:method}. Following \cite{wu2017rgb}, the two-stream backbone is modified from ResNet-18, with its first layer altered to be the modality-specific convolution layers, and the subsequent layers serving as shared convolution layers. For silhouettes, we define the processing as follows:
\begin{equation}
    F_{I} = L_{S}(I),
\end{equation}
where $I$ represents the silhouette and $L_{S}$ denotes the silhouette-specific convolution layer, which produces the feature map $F_{I}$.

Considering the modality differences and the need for consistency in data processing, we first project and interpolate point clouds to generate depths with the same resolution as silhouettes. And the processing is as follows: 
\begin{equation}
    F_{P} = L_{P}(\text{PI}(P)),
\end{equation}
where $P$ denotes the point cloud, $\text{PI}$ represents the projection and interpolation process, and $L_{P}$ is the depth-specific convolution layer, which produces the feature map $F_{P}$ for the point cloud.

After extracting modality-specific features $F_P$ and $F_S$, we aim to unify these features within a consistent feature space. To achieve this, we employ a shared convolutional layer, denoted as $L_{\text{sh}}$, to process both $F_P$ and $F_S$:
\begin{equation}
    F' = L_{\text{sh}}(F_m), \text{ where } F_m = F_P \text{ or } F_S,
\end{equation}
where $F'$ is the output feature map of $L_{\text{sh}}$.

\noindent\textbf{Temporal and Spatial Pooling.}
In our cross-modality network, we follow commonly used strategies in gait recognition for temporal aggregation and spatial aggregation, specifically through temporal pooling and horizontal pooling \cite{fan2023opengait}. These techniques are essential for effectively summarizing and interpreting the dynamic and spatial aspects of gait data. Horizontal pooling divides the input feature map horizontally into $N_H$ parts, with each part being aggregated into a single feature vector:
\begin{equation}
\text{HP} : \mathbb{R}^{T \times H \times W \times C} \to \mathbb{R}^{T \times N_H \times C},
\end{equation}
where $T$ denotes the sequence length, $H$ and $W$ represent the height and width of the feature map, respectively, and $C$ indicates the dimension of the features.

\noindent Temporal pooling allows for the aggregation of features from variable-length sequences, enabling the model to capture the final sequence-level gait representation:
\begin{equation}
\text{TP} : \mathbb{R}^{T \times N_H \times C} \to \mathbb{R}^{N_H \times C}.
\end{equation}

\noindent Therefore, we obtain $N_H$ feature vectors, which are further mapped to the metric space using $N_H$ independent Multi-Layer Perceptrons. These mapped vectors are then concatenated to form the final sequence-level feature embedding. 

\noindent\textbf{Training and Inference.}
The cross-modality network of CL-Gait is trained with cross-modality triplet loss and cross-entropy loss. We have modified the triplet loss to accommodate cross-modality feature learning, where each modality is alternately used as the anchor, while the corresponding positive and negative samples are selected from the other modality, formulated as:
\begin{equation}
    \mathcal{L}_{\text{cross-triplet}} = \frac{1}{2} \left( \mathcal{L}_{\text{triplet}}(P, I_{\text{pos}}, I_{\text{neg}}) + \mathcal{L}_{\text{triplet}}(I, P_{\text{pos}}, P_{\text{neg}}) \right).
\end{equation}
The training loss is a combination of both losses with a hyperparameter $\gamma$:
\begin{equation}
	\mathcal{L} = \mathcal{L}_{\text{cross-triplet}} + \gamma \mathcal{L}_{\text{ce}},
\end{equation}

\noindent where $\mathcal{L}_{\text{ce}}$ is the cross-entropy loss. During inference, the similarity between samples from different modalities is measured using the Euclidean distance.

\subsection{Contrastive Silhouette-point Pre-training}
Based on our observations in gait recognition tasks, the significant modality difference between 3D point clouds and 2D images could be a critical factor affecting model performance. To be specific, point clouds focus more on the 3D positioning of body parts, whereas images concentrate on the contour information of the individual. It's crucial to establish connections between the distinct information focused on by each modality. Inspired by CLIP \cite{radford2021learning}, we propose a contrastive silhouette-point pre-training (CSPP) strategy to align the feature spaces of both modalities in the convolution-based encoders, as shown in Fig. \ref{fig:method_pre}. 

The pre-training process does not require identity labels from the samples for supervision. Trained on paired single-view data from aligned camera and LiDAR, the pre-training could make the model focus on learning a robust representation that bridges the gap between the modalities without direct identity-based guidance, and enhance the performance of the cross-modality network.


\noindent\textbf{Pre-training Process.} For pre-training, CL-Gait's input shifts from sequence data to numerous pairs of 3D point clouds and 2D silhouettes. These pairs are defined as data collected from the same individual at the same moment, encompassing both modalities. This approach ensures that the network learns to reconcile the inherent differences between the 3D and 2D representations, focusing on the alignment of features that accurately reflect the same gait patterns across modalities. 

The paired data is processed by the two-stream backbone and spatial pooling operations to obtain paired feature embeddings, i.e., $\mathcal{S} = \{S_{i}|i=1, 2..., N\}$ for silhouettes and $\mathcal{P} = \{P_{i}|i=1, 2..., N\}$ for point clouds. They are then utilized to compute the loss for contrastive representation learning. 

\noindent\textbf{Contrastive Learning loss.} The cornerstone of our pre-training is the contrastive learning loss \cite{radford2021learning}. Given the paired feature embeddings $\mathcal{S} = \{S_{i}|i=1, 2..., N\}$ and $\mathcal{P} = \{P_{i}|i=1, 2..., N\}$, there are $N \times N$ possible (silhouettes, point clouds) pairings. CL-Gait is trained to learn a multi-modality embedding space by maxmizing the cosine similarity of the feature embeddings of the $N$ real pairs while minimizing the cosine similarity of the embeddings of the $N^2 - N$ incorrect pairings. The cosine similarity matrix is calculated as follows:
\begin{equation}
    \mathcal{M} = N_{\text{L2}}(\mathcal{S}) \times N_{\text{L2}}(\mathcal{P})^T, 
\end{equation}
where $\mathcal{S} \in \mathbb{R}^{N \times D}$, $\mathcal{P} \in \mathbb{R}^{N \times D}$, $\mathcal{M} \in \mathbb{R}^{N \times N}$, $N_{\text{L2}}(\cdot)$ denotes L2 normalization, and $(\cdot) ^T$ denotes the transpose of the matrix. Then, a symmetric cross entropy loss over these similarity scores is optimized:
\begin{equation}
    \mathcal{L}_{\text{con}} = \frac{1}{2} \left( \mathcal{L}_{\text{ce}}(\mathcal{M}, G) + \mathcal{L}_{\text{ce}}(\mathcal{M}^T, G) \right),
\end{equation}
where $G = \{1, 2..., N\}$. $G$ represents the labels used for calculating the cross-entropy loss, indicating that pairs corresponding to the similarities on the diagonal of matrices $\mathcal{M}$ and $\mathcal{M}^T$ are considered positive samples, while all others are treated as negative samples. This approach ensures that the network learns to distinguish between matching and non-matching modality pairs effectively.

In the process described above, each frame of point cloud or silhouette, after passing through the pre-trained network, yields a feature embedding of size $(1, C)$, indicating global pooling is applied to the output feature map of the backbone. For practical application and to align each local feature embedding, horizontal pooling is used to obtain features of size $(N, C)$ for each frame. During the computation of the contrastive learning loss, pairs of features from the same individual, moment, and part across two modalities are considered positive samples, with all others being negative. Through our practice, we have found that adopting a strategy of aligning local features during the pre-training phase is more suitable for gait recognition tasks, leading to performance improvements.  

\begin{figure*}[t]
	\centering
	\includegraphics[width=0.95\linewidth]{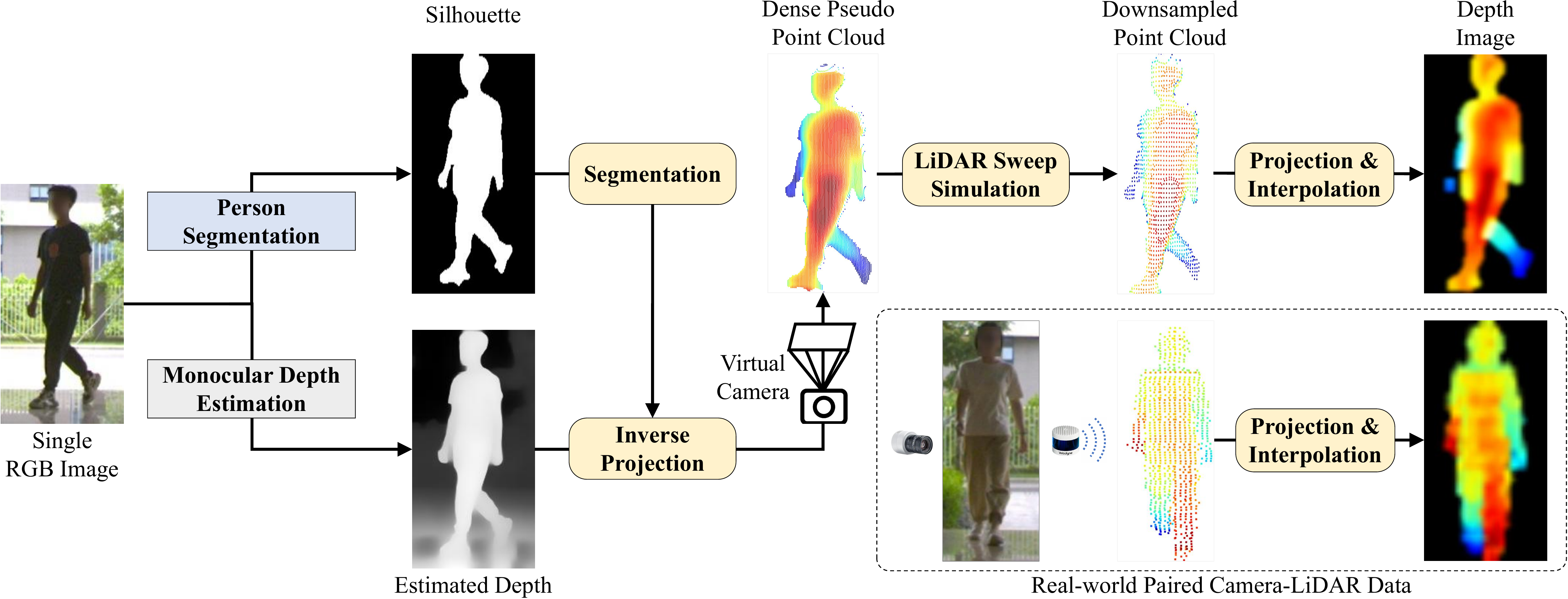}
	\caption{Illustration of paired gait data generation from single RGB images for contrastive pre-training. The quality of the synthesized data is comparable to the real-world data, making it possible to synthesize large-scale pre-training data.}
	\label{fig:cspp}
\end{figure*}

\subsection{Generation of Pre-training Gait Data}

Pre-training on real and large-scale data is challenging due to the high cost of acquiring paired RGB and point cloud data. To address this issue, we propose a method for generating pseudo data based on monocular depth estimation. As illustrated in Fig. \ref{fig:cspp}, we use Depth Anything \cite{yang2024depth} to estimate dense depth $D \in \mathbb{R}^{H\times W}$ from large-scale single RGB images. Then, utilizing a virtual camera with intrinsic $K$:
\begin{equation}
K = 
\begin{bmatrix}
f_x & 0 & c_x \\
0 & f_y & c_y \\
0 & 0 & 1
\end{bmatrix},
\end{equation}
where $f_x$ and $f_x$ are the focal lengths, and $c_x$ and $c_x$ are the coordinates of the principal point. Then the estimated depth is mapped to pseudo point clouds:
\begin{equation}
    [P', 1]^{T} = K^{-1}  D  [u, v, 1]^{T},
\end{equation}
where $P' \in \mathbb{R}^{H\times W \times 3}$ is the point coordinates, $K^{-1}$ indicates the inverse of intrinsic $K$, and $u, v$ are the image pixel coordinates. The pseudo points $P'$, after being downsampled by voxel grid into $P'_{down}$, can be used for contrastive pre-training between camera silhouettes and LiDAR point clouds. To be specific, the downsampled points $P'_{down}$ are projected back into the image pixel coordinate to obtain the depth images for contrastive pre-training. Compared to point clouds, RGB images are less costly to collect and easier to acquire, with many public datasets of pedestrian image already available. Our proposed method makes it feasible to synthesize large-scale data for contrastive pre-training.

\section{Experiments}

\subsection{Datasets}

\noindent\textbf{The SUSTech1K Dataset} \cite{shen2023lidargait} is the first large-scale LiDAR-based gait recognition dataset collected by a LiDAR sensor and an RGB camera. The dataset contains $25,239$ sequences from $1,050$ subjects and covers many variations, including visibility, views, occlusions, clothing, carrying, and scenes. It is timestamped frames for each modality of frames. The first single-modality 3D gait recognition framework is trained on SUSTech1K using point clouds. In our study, we explore new uses for the SUSTech1K dataset. Our approach involves both images and point clouds, which marks the first attempt of cross-modality gait recognition with this dataset. We also introduce a novel pre-training strategy based on paired images and point clouds from the dataset. This strategy utilizes contrastive learning for cross-modality matching between images and point clouds and improves the performance of our proposed cross-modality gait method.

\noindent\textbf{The HITSZ-VCM Dataset and LIP Dataset} are large-scale datasets, which contains a large amount of person videos/images captured by cameras. In detail, the HITSZ-VCM dataset \cite{lin2022learning} contains $927$ valid identities with image sequences. The LIP dataset \cite{gong2017look} contains about $50,000$ identities but with only one image for each. Totally, there are $251,452$ RGB images in the HITSZ-VCM dataset and $50,000$ in the LIP dataset. We implement the pre-training strategy on these datasets containing only RGB images, to explore the impact of pre-training with out-of-domain data on this task. Given that there is no point cloud data paired with these images, we use monocular depth estimation to generate pseudo point clouds, as shown in Fig.~\ref{fig:cspp}. This greatly expands the amount of data available for pre-training at scale. This also improves the accuracy of cross-modality recognition between 2D and 3D spaces, which marks the effectiveness of our proposed contrastive pre-training strategy.

\subsection{Experimental Setup}

\noindent\textbf{Implementation Details.} Following \cite{shen2023lidargait}, all the camera-based silhouettes and LiDAR-based depth images are aligned and then resized into the resolution of $64 × 64$. The total number of iterations is set to $120,000$ for pre-training and $60,000$ for fine-tuning. The Adam optimizer is used to prevent the issue of gradient vanishing because of low-quality silhouettes. The triplet and cross-entropy loss weights are set to the same. All comparison methods are trained using the same training strategy as LidarGait \cite{shen2023lidargait}. The OpenGait codebase \cite{fan2023opengait} is utilized to conduct all experiments.
  
\noindent\textbf{Evaluation Protocol.} All experiments are conducted on SUSTech1K. Following LidarGait \cite{shen2023lidargait}, the dataset is divided into three splits: a training set with $250$ identities and $6,011$ sequences, a validation set with $6,025$ sequences from $250$ unseen identities, and a test set with the remaining $550$ identities and $13,203$ sequences. The SUSTech1K dataset provides gait sequences from multiple viewpoints. Thus, we also adopt the cross-view evaluation protocol \cite{takemura2018multi,yu2006framework} used in CASIA-B and OUMVLP. During the test, the sequences in normal conditions are grouped into gallery sets, and the sequences in variant conditions are taken as probe sets. To accurately assess the performance of camera-LiDAR cross-modality gait recognition, we evaluate the results using point cloud data as the probe and silhouette data as the gallery, i.e., LiDAR to Camera (L to C). We also evaluate the results using silhouette data as the probe and point cloud data as the gallery, i.e., Camera to LiDAR (C to L).

\subsection{Comparative Methods} 
We evaluate several commonly used structures for cross-modality modeling as baselines, which basically follows the setting of \cite{wu2017rgb}. The baseline models includes one-steam structure, asymmetric FC layer structure, image-point structure, and the proposed two-stream structure. We apply residual block in ResNet-18 \cite{he2016deep} as the base convolution block for all the four structures. For them, the loss function are weighted cross-entropy loss and triplet loss \cite{chao2019gaitset}, which is commonly used and relatively stable. And all of the hyper parameters are kept the same. As for the input of one-steam structure, asymmetric FC layer structure and two-stream structure, the point clouds are converted to three-channel colored depth images and adjusted to the resolution of $64\times64$ with zero padding. We introduce the four structures as following:

\noindent\textbf{One-stream Structure} is the most
commonly used in vision tasks, where all parameters are shared in the whole network. Therefore, we organize silhouettes and depth images obtained from point clouds into the same dimensions before feeding them into the network. 

\noindent\textbf{Asymmetric FC Layer Structure} is used in multi-domain tasks, for example, IDR \cite{he2017learning} for VIS-NIR face recognition. This structure shares nearly all parameters except for the last FC layer.
It assumes that the feature extraction for different modalities can be same and adaptation is achieved in the backbone.

\noindent\textbf{Image-point Structure} directly uses point clouds as input, rather than projected depth images. It employs two types of encoders: an image encoder for the silhouette input and a point cloud encoder for the point cloud input. Under this structure, we implement three commonly used point cloud encoders: GCN \cite{guo2023lidar}, PointNet \cite{qi2017pointnet}, and Point-Transformer \cite{zhao2021point}.

\noindent\textbf{Two-stream Structure} is commonly used in cross-modality matching tasks. It utilizes modality-specific modules in the shallow layers of the network and uniform modules with shared parameters in the deeper layers. Based on the two-stream structure, our proposed CL-Gait is equipped with a horizontal pooling module (HP) and a contrastive silhouette-point pre-training strategy (CSPP). To show the effectiveness of the HP module and CSPP strategy of CL-Gait, we implement three additional networks: 1) a two-stream backbone without CSPP, using global pooling instead of HP; 2) a two-stream backbone without CSPP, using HP; 3) a two-stream backbone with CSPP, using global pooling.
\begin{table}[t]
\centering
\caption{Evaluation with different structures on SUSTech1K valid + test set. We use ResNet-18 to extract image features by default. ``L to C'' indicates the results with point clouds as probe and silhouettes as gallery, and ``C to L'' indicates the reverse.}
\vspace{-.2cm}
\label{my-label}
\resizebox{\textwidth}{!}{%
\begin{scriptsize}
\begin{threeparttable}
\begin{tabular}{@{}ll|ccc|ccc@{}}
\toprule
\multirow{2}{*}{\textbf{Structure}} & \multirow{2}{*}{\textbf{Model}} & \multicolumn{3}{c|}{\textbf{L to C}} & \multicolumn{3}{c}{\textbf{C to L}} \\
~ & ~ & \textbf{Rank-1} & \textbf{Rank-3} & \textbf{Rank-5} & \textbf{Rank-1} & \textbf{Rank-3} & \textbf{Rank-5} \\ \midrule
 One-stream & ResNet & 2.60 & 5.61 & 7.99 & 14.94 & 25.92 & 31.87  \\ \hline
 Asymmetric FC & ResNet & 3.08 & 7.23 & 10.52 & 16.43 & 29.13 & 35.90  \\ \hline
 \multirow{3}{*}{Image-point} & GCN \cite{guo2023lidar} & 12.20 & 25.01 & 33.23 & 14.30 & 28.38 & 36.62  \\
 & PointNet \cite{qi2017pointnet} & 7.11 & 16.61 & 23.35 & 8.40 & 19.25 & 26.90  \\
 & Point-Transformer \cite{zhao2021point} & 13.25 & 26.72 & 35.29 & 15.42 & 30.04 & 38.70 \\ \hline
 \multirow{4}{*}{Two-stream} & ResNet & 30.39 & 48.77 & 57.31 & 35.26 & 54.42 & 62.84  \\ 
 & ResNet+CSPP & 32.18 & 50.85 & 59.22 & 36.54 & 55.75 & 64.11  \\
 & ResNet+HP & 45.41 & 62.89 & 69.71 & 42.30 & 59.67 & 67.12  \\
 & CLGait (ours)\tnote{*} & \textbf{53.29} & \textbf{69.54} & \textbf{75.59} & \textbf{55.12} & \textbf{71.23} & \textbf{77.31}  \\ \bottomrule
\end{tabular}
\begin{tablenotes}
    \footnotesize 
    \item[*] i.e., ResNet+CSPP+HP.
\end{tablenotes}
\end{threeparttable}
\end{scriptsize}
}
\label{tab:main-table}
\end{table}

\subsection{Comparative Results}
Following the cross-view and cross-modality evaluation protocol, we report the comparative results of the methods above in the Tab.~\ref{tab:main-table}. From the results, we can obtain the following observations: 1) CL-Gait demonstrates its superiority to all existing structures and methods, primarily due to the feature extraction capabilities of the two-stream structure. Additionally, the use of pre-training effectively mitigates modality discrepancy, further improving its performance. 2) Experiments under the two-stream structure can be considered as an ablation study, highlighting the important roles of CSPP and HP. Adding HP to the ResNet base improves the average rank-1 by approximately $11\%$, and further incorporating CSPP increases the average rank-1 by about another $10\%$. 3) The performance of the one-stream structure is poor, primarily because the one-stream structure uses an identical network to process both silhouettes and point clouds, failing to address the modality discrepancy between them. 4) Although the Asymmetric FC structure employs different FC layers for the two modalities, its performance remains poor. This indicates that modality-specific adaptations are best applied at the shallower levels of the network. Furthermore, during the training process, the asymmetric FC layers tend to cause the network to overfit quickly. 5) All methods utilizing the two-stream structure outperform those based on the image-point structure. This likely results from the depth images, obtained through projection, having data consistency and smaller modality discrepancy with silhouettes compared to point cloud inputs.

\begin{figure*}[t]
	\centering
	\includegraphics[width=0.9\linewidth]{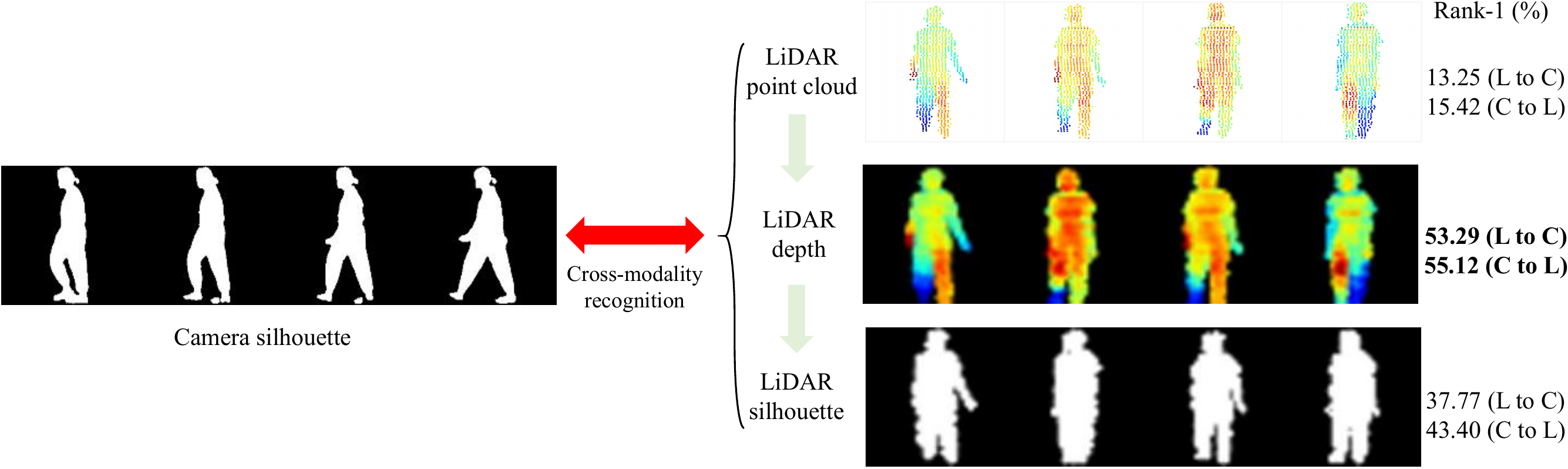}
     \vspace{-.1cm}
	\caption{Comparison of different input forms of LiDAR data. The results show that the projected and interpolated depth from point cloud works best for the cross-modality matching. This indicates that the 3D geometry information are essential. For each input form, the best-performing model is utilized.}
	\label{fig:3d-info}
	\vspace{-.2cm}
\end{figure*}

\subsection{Discussion}
\noindent\textbf{3D Geometry Information.} To evaluate the effectiveness of depth information for cross-modality gait recognition, we compare another type of data as input for the second modality, e.g., LiDAR silhouettes. LiDAR silhouettes are obtained by range-view projection of point cloud sets and contains no 3D geometry information. The results are shown in Fig. \ref{fig:3d-info}. When 3D geometry information is not included, the performance of LiDAR silhouettes is much lower. This is because the camera has a much higher resolution than LiDAR, so the silhouettes from the camera can have more details. 
Integrating depth information can improve average rank-1 accuracy from $40.59\%$ to $54.21\%$, validating the necessity and effectiveness of 3D information for cross-modality gait recognition. One important reason may be that the 3D information contained in the point cloud can compensate for the difference in viewing angles to some extent. Furthermore, we include the performance of direct point cloud input in Fig. \ref{fig:3d-info} to compare with depth images and illustrate the input form of point clouds.
 
\noindent\textbf{Contrastive Pre-training Datasets.} To make up for the absence of paired camera-LiDAR data, we further propose a large-scale data generation strategy. Camera silhouette and LiDAR point data are generated from real-world RGB images for contrastive pre-training on such paired pseudo data. To valid its performance, we construct training sets from two different datasets, including LIP and HITSZ-VCM. These two datasets are commonly used for camera-based person segmentation and re-identification and contain single still RGB images. To ensure fair comparisons, we also conduct experiments on SUSTech1K training set, which contains real paired point clouds and camera silhouettes.

As shown in Tab. \ref{tab:sub-table}, when the amount of data is similar using data generated from the LIP dataset, the performance of pre-training on generated pseudo data is comparable to that of real data. When pre-trained on a larger scale of synthetic pseudo data generated from the HITSZ-VSM dataset, the accuracy from LiDAR to camera improves and exceeds that on the real data, proving the effectiveness of our data generation strategy for contrastive pre-training. However, the accuracy from camera to LiDAR decreases a little, probably due to the difference between generated point clouds from estimated depth and real points from LiDAR.

\begin{table}[t]
\centering
\caption{Performance with different pre-training datasets. Using estimated depth for pre-training obtains close results compared with using real paired point-image dataset.}
\vspace{-.2cm}
\label{my-label}
\begin{scriptsize}
\begin{tabular}{@{}lc|ccc|ccc@{}}
\toprule
\multirow{2}{*}{\textbf{Dataset}} & \textbf{Data} & \multicolumn{3}{c|}{\textbf{L to C}} & \multicolumn{3}{c}{\textbf{C to L}} \\
~ & \textbf{Amount} & \textbf{Rank-1} & \textbf{Rank-3} & \textbf{Rank-5} & \textbf{Rank-1} & \textbf{Rank-3} & \textbf{Rank-5} \\ \midrule
 SUSTech1K & 120K & 51.60 & 68.22 & 74.43 & \textbf{56.12} & \textbf{72.15} & \textbf{77.95}  \\ \hline
 LIP & 50K & 51.00 & 68.09 & 73.99 & 55.07 & 71.59 & 77.59  \\ 
 HITSZ-VCM & 770K & \textbf{53.29} & \textbf{69.54} & \textbf{75.59} & 55.12 & 71.23 & 77.31  \\ \bottomrule
\end{tabular}
\end{scriptsize}
\label{tab:sub-table}
\end{table}

\begin{table}[t]
\centering
\caption{Performance of CL-Gait with different temporal modules.}
\vspace{-.2cm}
\begin{scriptsize}
\begin{tabular}{@{}l|ccc|ccc@{}}
\toprule
\multirow{2}{*}{\textbf{Feature}} & \multicolumn{3}{c|}{\textbf{L to C}} & \multicolumn{3}{c}{\textbf{C to L}} \\
~ & \textbf{Rank-1} & \textbf{Rank-3} & \textbf{Rank-5} & \textbf{Rank-1} & \textbf{Rank-3} & \textbf{Rank-5} \\ \midrule
Temporal Pooling \cite{fan2023opengait} & \textbf{53.29} & \textbf{69.54} & \textbf{75.59} & \textbf{55.12} & \textbf{71.23} & \textbf{77.31}  \\ 
LSTM \cite{hochreiter1997long} & 37.40 & 55.01 &60.89& 39.47 & 56.41 &  62.72 \\
Bi-LSTM \cite{schuster1997bidirectional} & 39.63 & 55.55 & 61.83 & 39.88 & 57.82 &63.71\\
Transformer \cite{vaswani2017attention} & 44.80 & 61.31 & 67.36 & 47.16 & 63.83 &69.11\\ \bottomrule
\end{tabular}
\end{scriptsize}
\label{tab:sub-tem}
\vspace{-.3cm}
\end{table}

\noindent\textbf{Temporal Modules.} In order to aggregate the different frames in the sequence, temporal modules are used to fuse the features of multiple frames. We evaluate the effectiveness of different temporal modules in combination with our proposed CL-Gait, as shown in Tab. \ref{tab:sub-tem}. We can observe that the simple temporal pooling structure demonstrates superiority over the other learning-based structure. This result indicates that temporal pooling can better fuse multi-frame features to match domain information more effectively. One reason may be that learning-based approaches such as LSTM and Transformer are more likely to overfit.




\section{Conclusion and Future work} 
This paper presents the first research on cross-modality gait recognition with point clouds and silhouettes from RGB images. Firstly, We propose a cross-modality gait recognition framework, named CL-Gait, that utilizes a two-stream network for feature embedding of different modalities, i.e., camera and LiDAR. Moreover, to mitigate modality discrepancy, we propose a contrastive pre-training strategy along with a large-scale data generation method. It can generate large number of data pairs (silhouette and point cloud) based on monocular depth estimation and pre-train the model in a contrastive manner.
Extensive experiments demonstrate that CL-Gait achieves commendable performance in cross-modality evaluation modes, proving the importance of maintaining modality consistency and the effectiveness of contrastive pre-training. This also highlights the potential of cross-modality gait recognition.

CL-Gait has shown remarkable results, but we believe there remains significant room for improvement. The modality and resolution differences between point clouds and silhouettes, along with the inherent motion blur in point clouds, may be key factors impacting model performance. 
Therefore, densifying point clouds or developing point cloud encoders specifically for cross-modality gait recognition could enhance model performance.

\section*{\centering \Large Camera-LiDAR Cross-modality Gait Recognition (Supplementary Material)}
\appendix

We provide more details in this supplementary material, including: 1) Experiments on the impact of modality on gait recognition. 2) Feature visualization about our contrastive silhouette-point pre-training strategy (CSPP). 3) Examples of generated multimodal gait data for contrastive pre-training. 

\section{Impact of Modality}

\begin{figure}
	\centering
        \vspace{-.2cm}
	\includegraphics[width=0.9\linewidth]{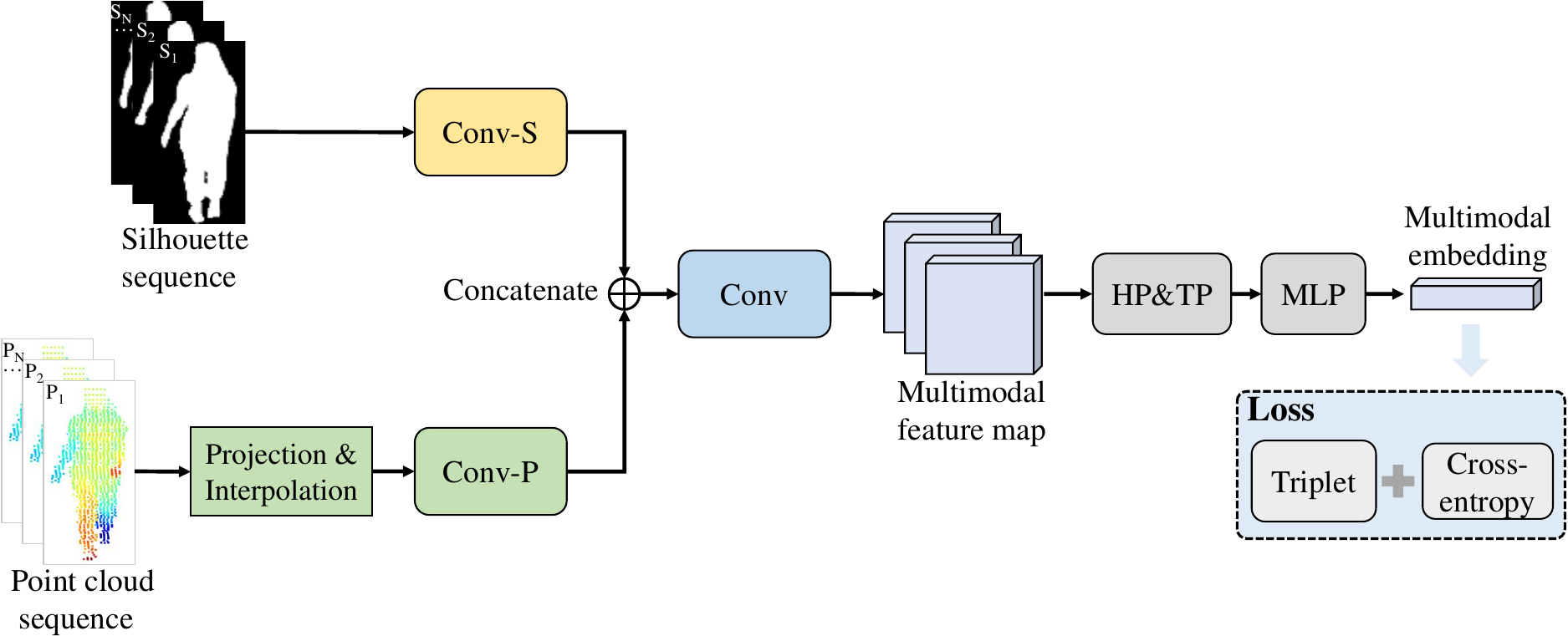}
        \vspace{-.2cm}
	\caption{The overview of CL-Gait-F. We modify the cross-modality network of CL-Gait to obtain CL-Gait-F for camera-LiDAR multi-modality gait recognition. The input to CL-Gait-F consists of synchronized sequences of silhouettes and point clouds.}
	\label{fig:method_fusion}
	
\end{figure}

To investigate the impact of different modalities on gait recognition tasks, we compare the performance of single-modality, multi-modality, and cross-modality gait recognition approaches. For single-modality methods, we compare with the state-of-the-art methods on the SUSTech1K dataset~\cite{shen2023lidargait}. 
Because there are no existing camera-LiDAR multi-modality methods for gait recognition, we modify the cross-modality network of CL-Gait to obtain CL-Gait-F, as illustrated in Fig.~\ref{fig:method_fusion}. The input to CL-Gait-F consists of synchronized sequences of silhouettes from camera and point clouds from LiDAR. The quantitative results are shown in Tab.~\ref{tab:modality}, from which the following observations can be obtained: 
1) The multi-modality method, CL-Gait-F, surpasses all single-modality methods, indicating the complementarity of the two modalities and also proving that the network of CL-Gait can effectively extract distinguishable features from both modalities. 2) Cross-modality gait recognition performs worse than other methods because it needs to handle the significant modality discrepancy between different modalities. It is a valuable but challenging task that still requires further research.

\begin{table}[]
\centering
\caption{Performance of state-of-the-art methods under different modalities on the SUSTech1K dataset. `-' indicates that the result is not reported in the paper.}
\label{tab:modality}
\setlength{\tabcolsep}{5pt}
\begin{footnotesize}
\begin{tabular}{cc|ccc}
\toprule
\textbf{Method} & \textbf{Modality} & \textbf{Rank-1} & \textbf{Rank-3} & \textbf{Rank-5} \\ \midrule
 GaitSet~\cite{chao2019gaitset} & \multirow{4}{*}{Camera} & 65.04 & - & 84.76 \\ 
 GaitPart~\cite{fan2020gaitpart} & ~ & 59.19 & - & 80.79 \\ 
 GaitGL~\cite{lin2022gaitgl} & ~ & 63.14 & - & 82.82 \\ 
 GaitBase~\cite{fan2023opengait} & ~ & 75.98 & 86.22 & 89.59 \\ \hline
 PointNet~\cite{qi2017pointnet} & \multirow{5}{*}{LiDAR}  & 31.33 & - & 59.75   \\ 
 PointNet++~\cite{qi2017pointnet++} & ~  & 50.78 & - & 82.38   \\ \
 PointTransformer~\cite{zhao2021point} & ~ & 44.37 & - & 76.70  \\ 
 SimpleView~\cite{goyal2021revisiting} & ~  & 64.83 & - & 85.77   \\ 
 LidarGait~\cite{shen2023lidargait} & ~  & 86.66 & 94.10 & 95.92 \\ \hline
 CL-Gait-F (ours) & Camera and LiDAR & \textbf{90.06} & \textbf{95.97} & \textbf{97.31} \\ \hline
 \multirow{2}{*}{CL-Gait (ours)} & LiDAR to Camera & 53.29 & 69.54 & 75.59 \\ 
 ~ & Camera to LiDAR & 55.12 & 71.23 & 77.31  \\ \bottomrule
\end{tabular}
\end{footnotesize}
\label{tab:sub-table}
\end{table}

\section{Feature Visualization}

\begin{figure}
	\centering
	\includegraphics[width=0.85\linewidth]{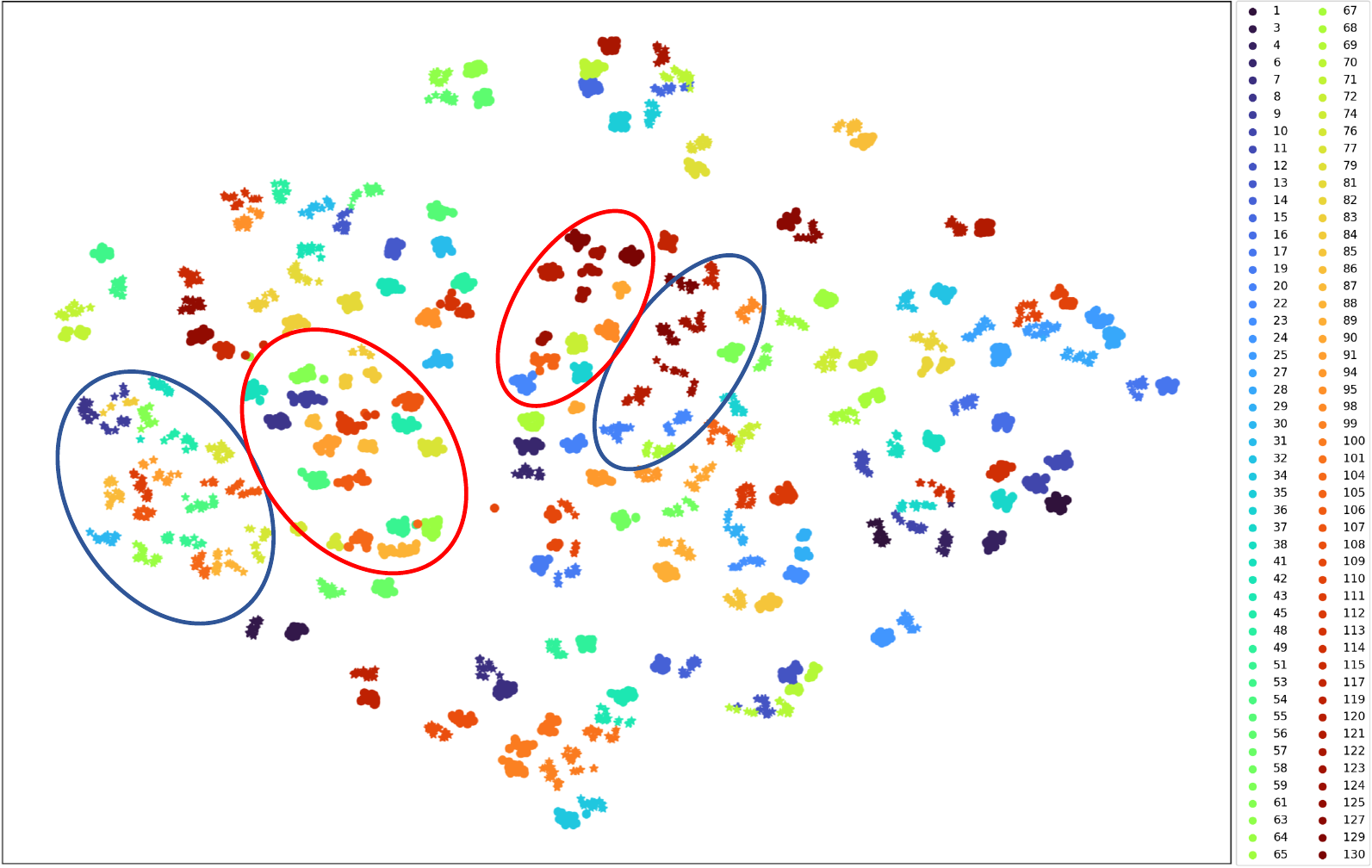}
	\caption{The feature distribution of the first 100 individuals in the SUSTech1K test set extracted by CL-Gait without pre-training (CL-Gait-B). Stars and points respectively represent the features of silhouette sequences and point cloud sequences, and distinct colors indicate different individuals. Features of the same modality for some individuals cluster together and are distant from features of another modality, as indicated by the elliptical circles. This is primarily due to the significant modality discrepancy between 2D silhouettes and 3D point clouds.}
	\label{fig:tsne_base}
	\vspace{-.2cm}
\end{figure}

\begin{figure}
	\centering
	\includegraphics[width=0.85\linewidth]{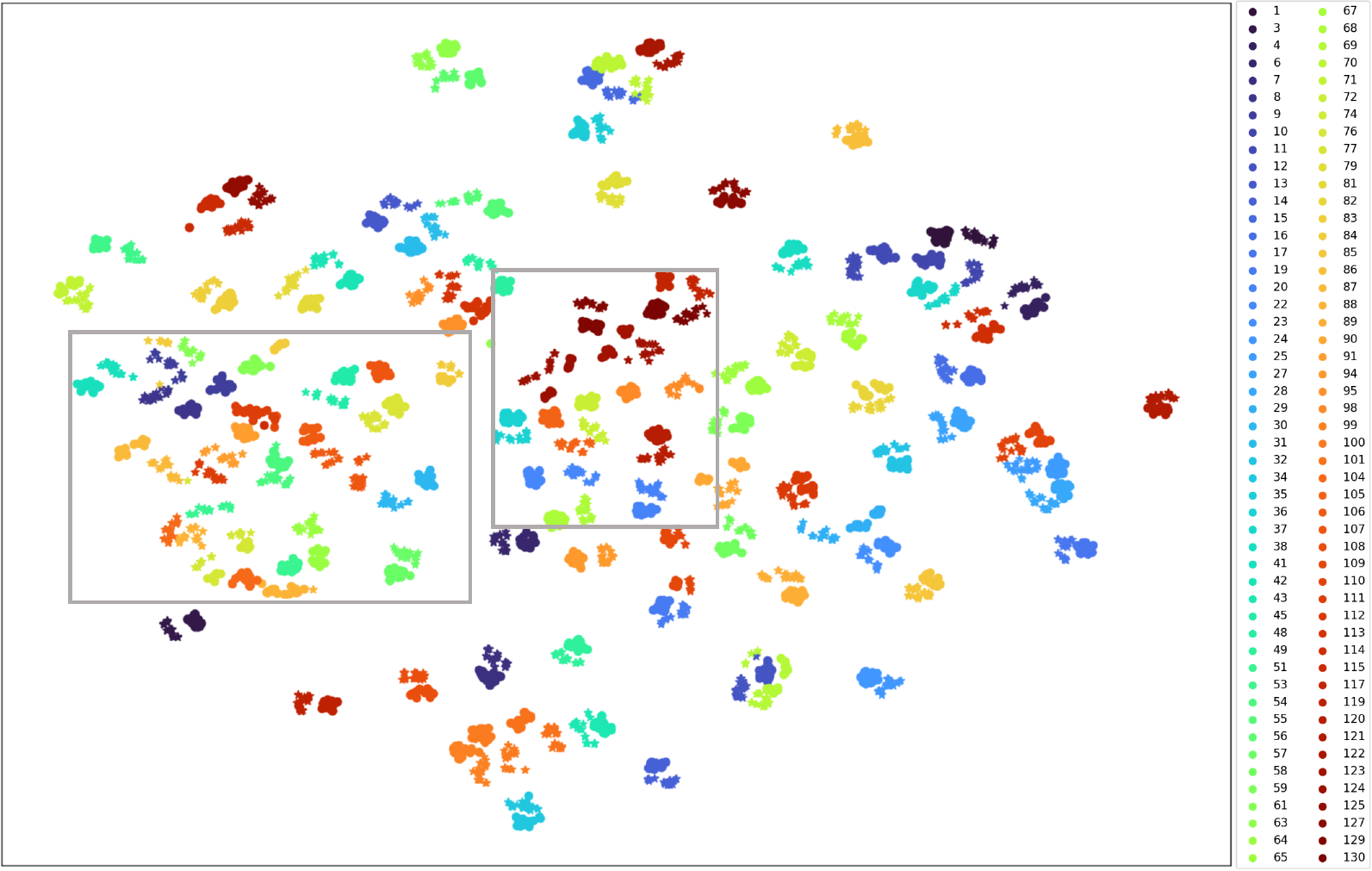}
	\caption{The feature distribution of the first 100 individuals in the SUSTech1K test set extracted by CL-Gait with contrastive pre-training. The individuals within gray boxes correspond to these within the elliptical circles in Fig.~\ref{fig:tsne_base}. The clustering phenomenon within the same modality has been greatly reduced, making cross-modality retrieval more accurate. This can be attributed to the effectiveness of contrastive pre-training in mitigating modality discrepancy.}
	\label{fig:tsne_pre}
\end{figure}

To visually demonstrate the effectiveness of our proposed contrastive silhouette-point pre-training strategy (CSPP), we use t-SNE to visualize the feature distributions of the first 100 individuals in the SUSTech1K test set extracted by CL-Gait with and without CSPP, as shown in Fig.~\ref{fig:tsne_base} and Fig.~\ref{fig:tsne_pre}. We refer to CL-Gait without pre-training as CL-Gait-B. We can observe that the feature distributions extracted by CL-Gait and CL-Gait-B share certain similarities. However, for cross-modality retrieval tasks, CL-Gait demonstrates better discriminative ability. In Fig.~\ref{fig:tsne_base}, features of the same modality for some individuals cluster together and are distant from features of another modality, as emphasized by the colored elliptical circles. This is primarily due to the significant modality discrepancy between 2D silhouettes and 3D point clouds. Conversely, in Fig.~\ref{fig:tsne_pre}, the clustering phenomenon within the same modality is significantly reduced, making cross-modality retrieval more accurate, as indicated by the gray boxes. We attribute this to the potent influence of our proposed CSPP, which effectively mitigates modality discrepancy.

\section{Generation of Pre-training Gait Data}
\subsection{Comparison with Real Gait Data}
\begin{figure*}[]
	\begin{center}
		\begin{subfigure}{0.46\linewidth}
			\begin{center}
				\includegraphics[width=\linewidth]{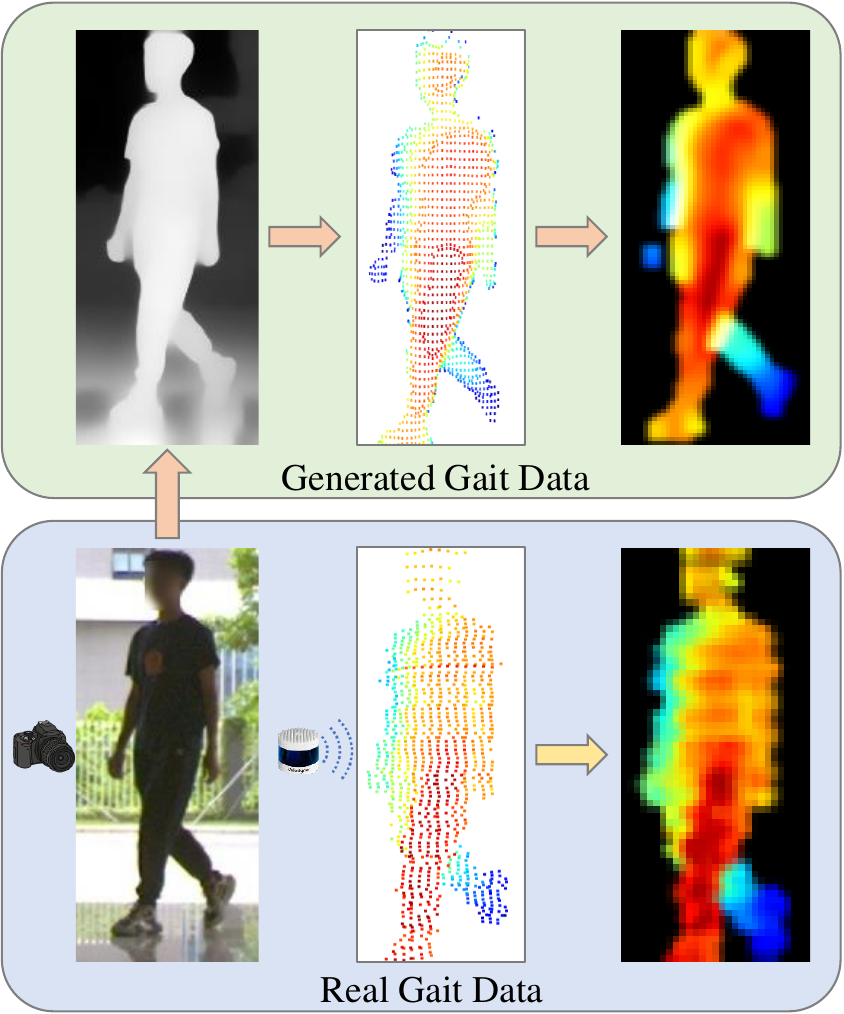}
				\caption{}\label{fig:gener1}
			\end{center}
		\end{subfigure}
		\quad\quad
		\begin{subfigure}{0.46\linewidth}
			\begin{center}
				\includegraphics[width=\linewidth]{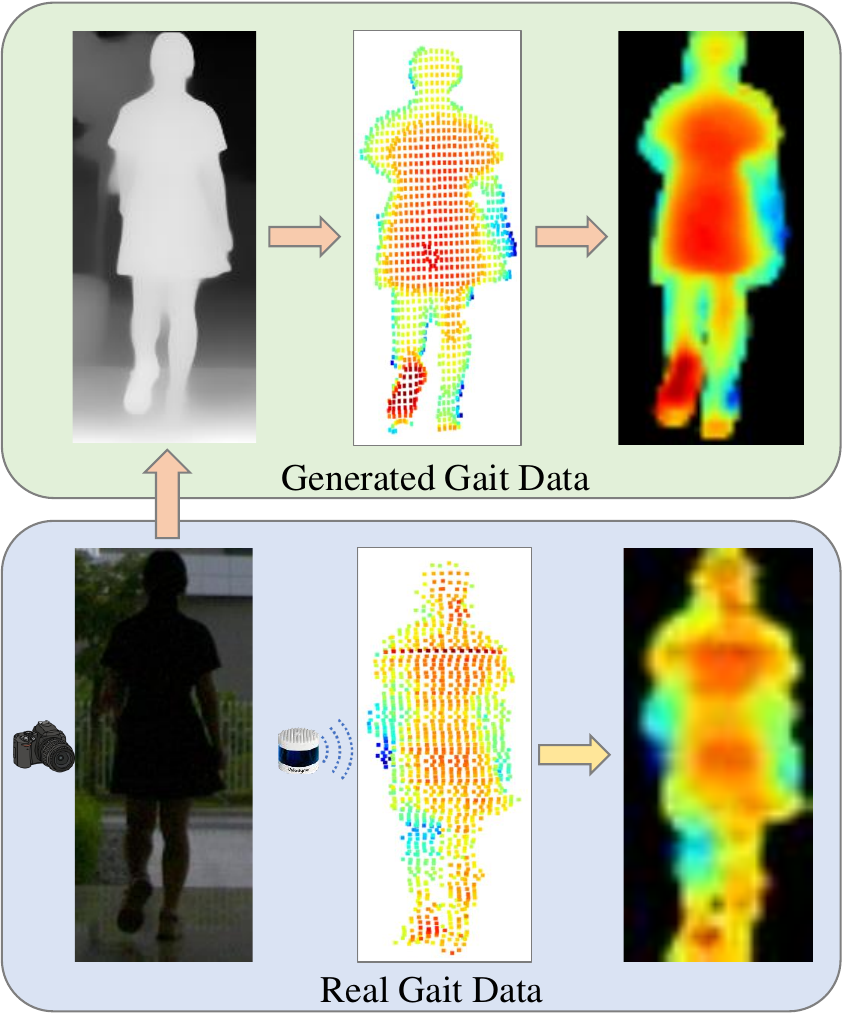}
				\caption{}
				\label{fig:gener2}
			\end{center}
		\end{subfigure}	
	\end{center}
	\vspace{-.2cm}
	\caption{Examples of generated multimodal gait data and corresponding real data. In each example, the green box contains the estimated and generated gait data, and the blue box contains the real data. Real data acquisition costs are high. In contrast, our synthetic data can be easily accessed at scale.}\label{fig:gener}
\end{figure*}

To demonstrate the effectiveness and realism of our proposed method of multimodal gait data generation, we present several generated examples and compare them with the real data, as shown in Fig.~\ref{fig:gener}. The examples are from SUSTech1K dataset, because it includes RGB images, real point clouds and corresponding depth images, for the comparison with our generated data. From Fig.~\ref{fig:gener}, we can observe that the depths estimated by our method are realistic. The generated point clouds and the depth maps obtained from point clouds are very close to the real data. Furthermore, as shown in Fig.~\ref{fig:gener2}, even in low-light environments, our method can still obtain accurate depth estimation and generate gait data that closely matches real-world conditions. The realism of the generated point clouds and depth maps, along with the consistency of the paired gait data, ensures the effective implementation of contrastive pre-training to mitigate modality discrepancy in cross-modality gait recognition tasks.

\subsection{Visualization of Generated Samples}
\begin{figure}[]
	\begin{center}
        
		\begin{subfigure}{0.87\linewidth}
          \setcounter{subfigure}{2}
			\begin{center}
				\includegraphics[width=\linewidth]{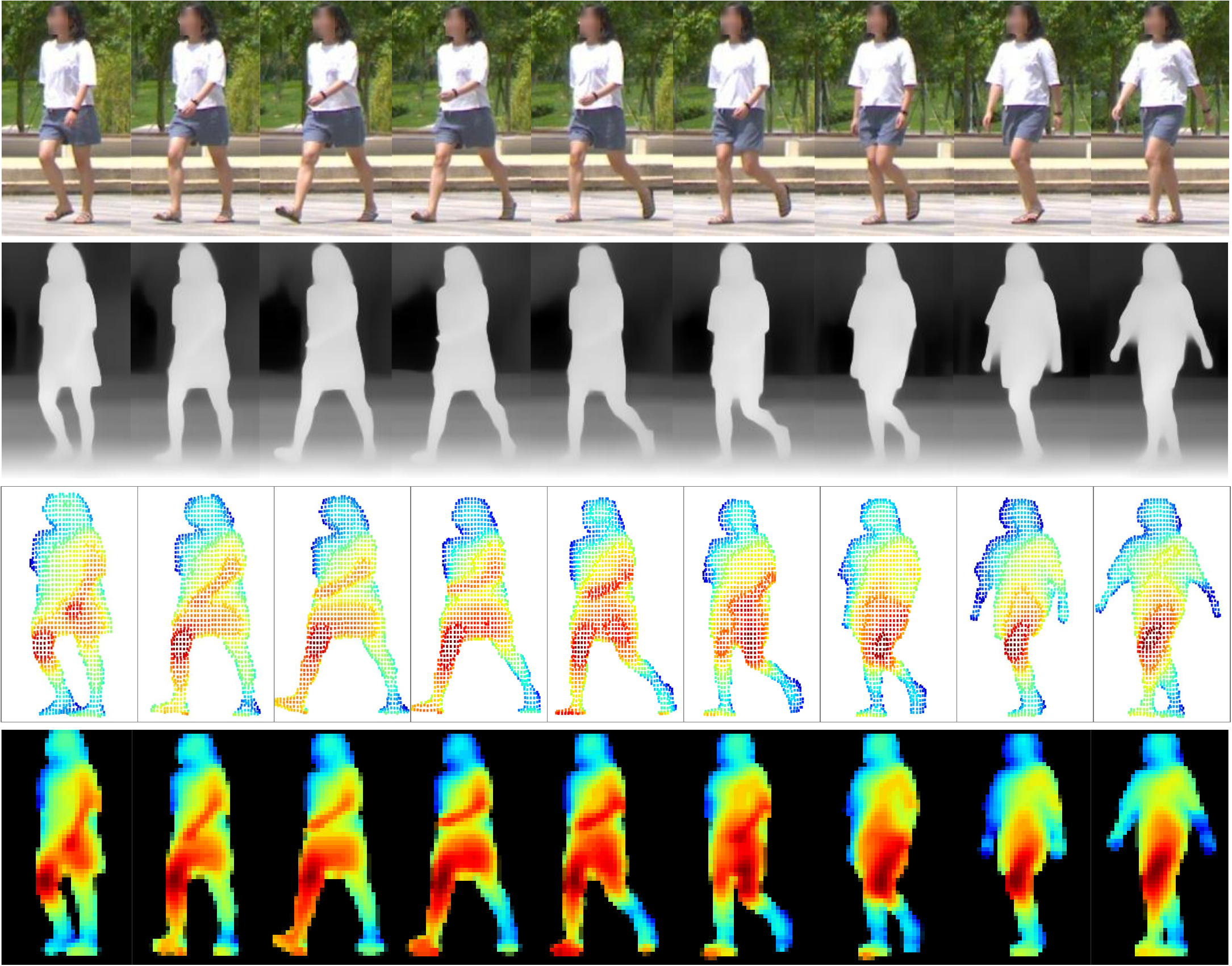}
				\caption{SUSTech1K~\cite{shen2023lidargait}, scene 3, view 45°.}\label{fig:gener2_3}
			\end{center}
		\end{subfigure}	
        \vskip 0.2cm
		\begin{subfigure}{0.87\linewidth}
			\begin{center}
				\includegraphics[width=\linewidth]{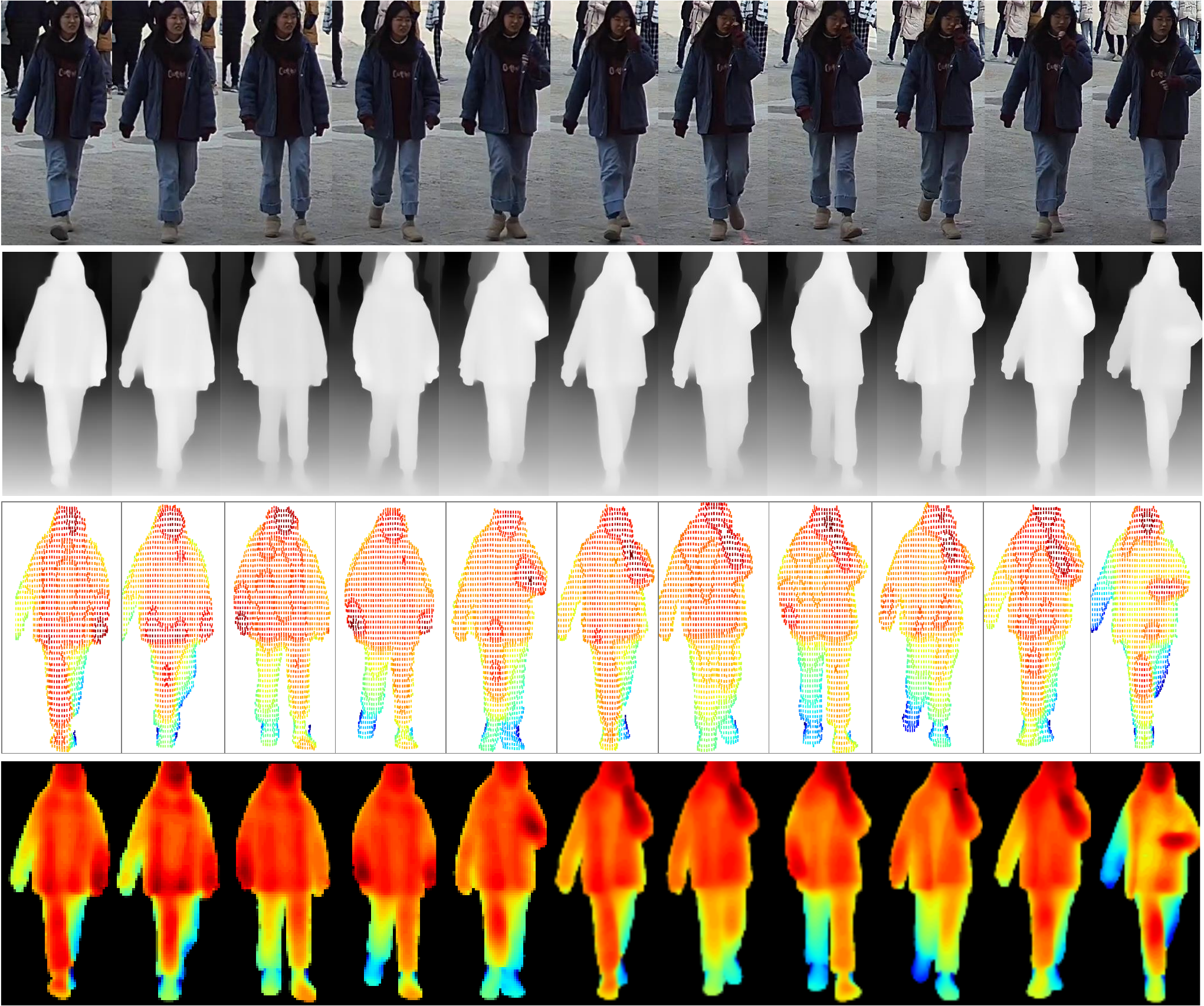}
				\caption{HITSZ-VCM~\cite{ling2020class}.}
				\label{fig:gener2_4}
			\end{center}
		\end{subfigure}	
	\end{center}
        \vspace{-.6cm}
	\caption{Examples of our multimodal gait data generation method. The examples are sequence data, covering individuals from different datasets, scenarios, and views.}\label{fig:gener_seque}
\end{figure}

Our CL-Gait can automatically generate training pairs from RGB images or videos without human involvement. Thus, huge amount of RGB images can be used to generate training pairs  for contrastive pre-training, which helps the supervised networks generalize to various scenes. Figure~\ref{fig:gener_seque} shows some generated samples from real-world video sequences that cover individuals from different datasets, scenarios, and views. These generated colored depth images, together with the person silhouettes, are then used as our training pairs for our proposed contrastive silhouette-point pre-training strategy.
\begin{figure}[]
	\begin{center}
		\begin{subfigure}{0.9\linewidth}
			\begin{center}
				\includegraphics[width=\linewidth]{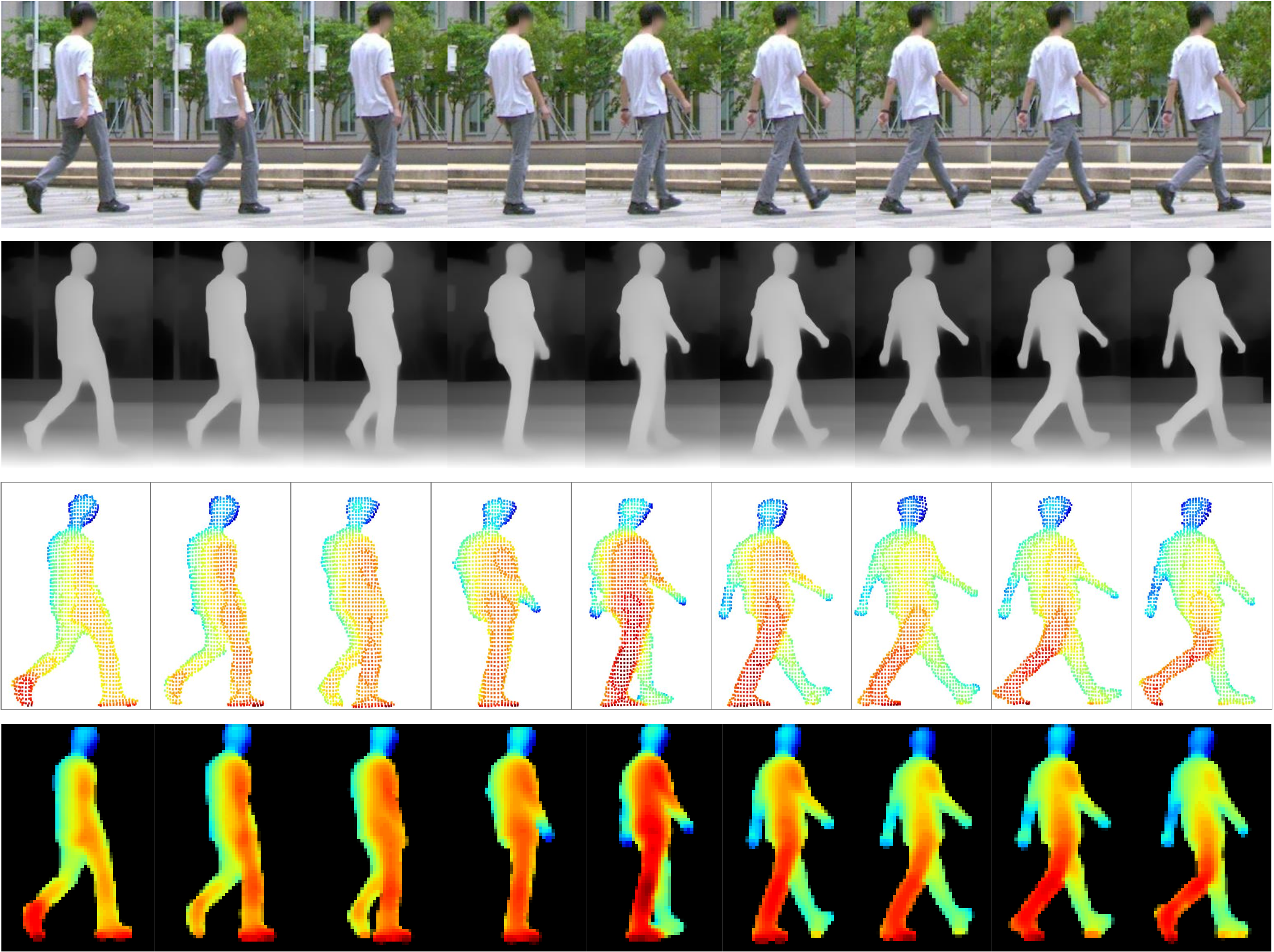}
				\caption{SUSTech1K~\cite{shen2023lidargait}, scene 1, view 225°.}\label{fig:gener2_1}
			\end{center}
		\end{subfigure} 
        \vskip 0.5cm
		\begin{subfigure}{0.9\linewidth}
			\begin{center}
				\includegraphics[width=\linewidth]{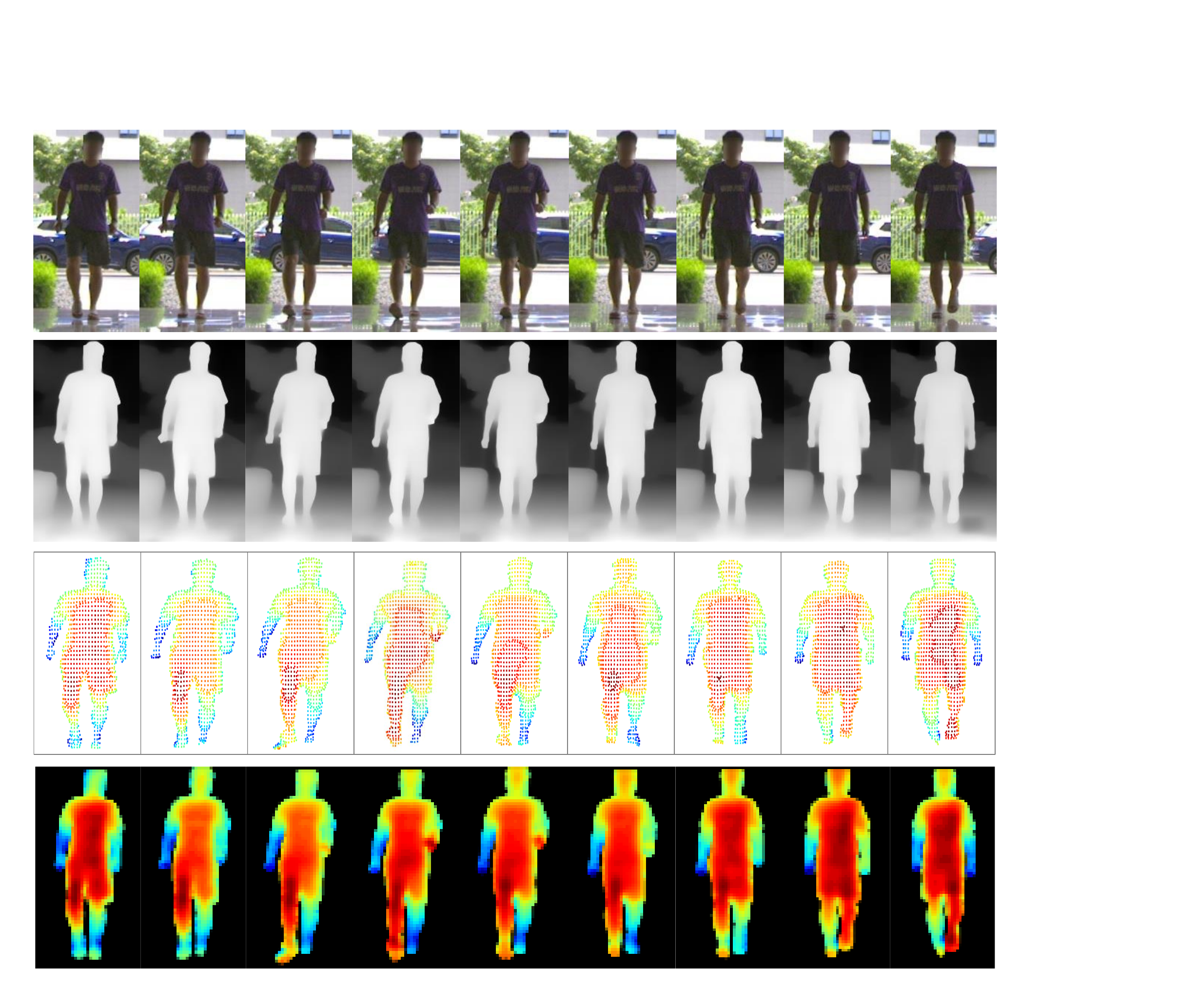}
				\caption{SUSTech1K~\cite{shen2023lidargait}, scene 2, view 0°.}
				\label{fig:gener2_2}
			\end{center}
		\end{subfigure}	
	\end{center}
	\label{fig:gener_seque_part1}
\end{figure}


%
%
\bibliographystyle{splncs04}
\bibliography{main}

\begin{thebibliography}{10}
\providecommand{\url}[1]{\texttt{#1}}
\providecommand{\urlprefix}{URL }
\providecommand{\doi}[1]{https://doi.org/#1}

\bibitem{bobick2001recognition}
Bobick, A.F., Davis, J.W.: The recognition of human movement using temporal templates. IEEE Transactions on Pattern Analysis and Machine Intelligence  \textbf{23}(3),  257--267 (2001)

\bibitem{chao2019gaitset}
Chao, H., He, Y., Zhang, J., Feng, J.: Gaitset: Regarding gait as a set for cross-view gait recognition. In: Proceedings of the AAAI Conference on Artificial Intelligence. vol.~33, pp. 8126--8133 (2019)

\bibitem{choi2020hi}
Choi, S., Lee, S., Kim, Y., Kim, T., Kim, C.: Hi-cmd: Hierarchical cross-modality disentanglement for visible-infrared person re-identification. In: Proceedings of the IEEE/CVF Conference on Computer Vision and Pattern Recognition. pp. 10257--10266 (2020)

\bibitem{dai2018cross}
Dai, P., Ji, R., Wang, H., Wu, Q., Huang, Y.: Cross-modality person re-identification with generative adversarial training. In: IJCAI. vol.~1, p.~6 (2018)

\bibitem{fan2023opengait}
Fan, C., Liang, J., Shen, C., Hou, S., Huang, Y., Yu, S.: Opengait: Revisiting gait recognition towards better practicality. In: Proceedings of the IEEE/CVF Conference on Computer Vision and Pattern Recognition. pp. 9707--9716 (2023)

\bibitem{fan2020gaitpart}
Fan, C., Peng, Y., Cao, C., Liu, X., Hou, S., Chi, J., Huang, Y., Li, Q., He, Z.: Gaitpart: Temporal part-based model for gait recognition. In: Proceedings of the IEEE/CVF Conference on Computer Vision and Pattern Recognition. pp. 14225--14233 (2020)

\bibitem{feng2019learning}
Feng, Z., Lai, J., Xie, X.: Learning modality-specific representations for visible-infrared person re-identification. IEEE Transactions on Image Processing  \textbf{29},  579--590 (2019)

\bibitem{gong2017look}
Gong, K., Liang, X., Zhang, D., Shen, X., Lin, L.: Look into person: Self-supervised structure-sensitive learning and a new benchmark for human parsing. In: Proceedings of the IEEE conference on computer vision and pattern recognition. pp. 932--940 (2017)

\bibitem{goyal2021revisiting}
Goyal, A., Law, H., Liu, B., Newell, A., Deng, J.: Revisiting point cloud shape classification with a simple and effective baseline. In: International Conference on Machine Learning. pp. 3809--3820. PMLR (2021)

\bibitem{guo2023lidar}
Guo, W., Pan, Z., Liang, Y., Xi, Z., Zhong, Z.C., Feng, J., el~al: Lidar-based person re-identification. arXiv preprint arXiv:2312.03033  (2023)

\bibitem{hao2019hsme}
Hao, Y., Wang, N., Li, J., Gao, X.: Hsme: Hypersphere manifold embedding for visible thermal person re-identification. In: Proceedings of the AAAI Conference on Artificial Intelligence. vol.~33, pp. 8385--8392 (2019)

\bibitem{he2016deep}
He, K., Zhang, X., Ren, S., Sun, J.: Deep residual learning for image recognition. In: Proceedings of the IEEE Conference on Computer Vision and Pattern Recognition. pp. 770--778 (2016)

\bibitem{he2017learning}
He, R., Wu, X., Sun, Z., Tan, T.: Learning invariant deep representation for nir-vis face recognition. In: Proceedings of the AAAI Conference on Artificial Intelligence. vol.~31 (2017)

\bibitem{hochreiter1997long}
Hochreiter, S., Schmidhuber, J.: Long short-term memory. Neural computation  \textbf{9}(8),  1735--1780 (1997)

\bibitem{jia2021scaling}
Jia, C., Yang, Y., Xia, Y., Chen, Y.T., Parekh, Z., Pham, H., el~al: Scaling up visual and vision-language representation learning with noisy text supervision. In: International Conference on Machine Learning. pp. 4904--4916. PMLR (2021)

\bibitem{kamath2021mdetr}
Kamath, A., Singh, M., LeCun, Y., Synnaeve, G., Misra, I., Carion, N.: Mdetr-modulated detection for end-to-end multi-modal understanding. In: Proceedings of the IEEE/CVF International Conference on Computer Vision. pp. 1780--1790 (2021)

\bibitem{li2020end}
Li, X., Makihara, Y., Xu, C., Yagi, Y., Yu, S., Ren, M.: End-to-end model-based gait recognition. In: Proceedings of the Asian Conference on Computer Vision (2020)

\bibitem{liang2022gaitedge}
Liang, J., Fan, C., Hou, S., Shen, C., Huang, Y., Yu, S.: Gaitedge: Beyond plain end-to-end gait recognition for better practicality. In: European Conference on Computer Vision. pp. 375--390. Springer (2022)

\bibitem{liao2020model}
Liao, R., Yu, S., An, W., Huang, Y.: A model-based gait recognition method with body pose and human prior knowledge. Pattern Recognition  \textbf{98},  107069 (2020)

\bibitem{lin2021gaitmask}
Lin, B., Liu, Y., Zhang, S.: Gaitmask: Mask-based model for gait recognition. In: BMVC. pp. 1--12 (2021)

\bibitem{lin2022gaitgl}
Lin, B., Zhang, S., Wang, M., Li, L., Yu, X.: Gaitgl: Learning discriminative global-local feature representations for gait recognition. arXiv preprint arXiv:2208.01380  (2022)

\bibitem{lin2022learning}
Lin, X., Li, J., Ma, Z., Li, H., Li, S., Xu, K., Lu, G., Zhang, D.: Learning modal-invariant and temporal-memory for video-based visible-infrared person re-identification. In: Proceedings of the IEEE/CVF Conference on Computer Vision and Pattern Recognition. pp. 20973--20982 (2022)

\bibitem{ling2020class}
Ling, Y., Zhong, Z., Luo, Z., Rota, P., Li, S., Sebe, N.: Class-aware modality mix and center-guided metric learning for visible-thermal person re-identification. In: Proceedings of the 28th ACM International Conference on Multimedia. pp. 889--897 (2020)

\bibitem{loper2023smpl}
Loper, M., Mahmood, N., Romero, J., Pons-Moll, G., Black, M.J.: Smpl: A skinned multi-person linear model. In: Seminal Graphics Papers: Pushing the Boundaries, Volume 2, pp. 851--866 (2023)

\bibitem{lu2020cross}
Lu, Y., Wu, Y., Liu, B., Zhang, T., Li, B., Chu, Q., Yu, N.: Cross-modality person re-identification with shared-specific feature transfer. In: Proceedings of the IEEE/CVF Conference on Computer Vision and Pattern Recognition. pp. 13379--13389 (2020)

\bibitem{nguyen2017person}
Nguyen, D.T., Hong, H.G., Kim, K.W., Park, K.R.: Person recognition system based on a combination of body images from visible light and thermal cameras. Sensors  \textbf{17}(3), ~605 (2017)

\bibitem{park2021learning}
Park, H., Lee, S., Lee, J., Ham, B.: Learning by aligning: Visible-infrared person re-identification using cross-modal correspondences. In: Proceedings of the IEEE/CVF International Conference on Computer Vision. pp. 12046--12055 (2021)

\bibitem{qi2017pointnet}
Qi, C.R., Su, H., Mo, K., Guibas, L.J.: Pointnet: Deep learning on point sets for {3D} classification and segmentation. In: Proceedings of the IEEE Conference on Computer Vision and Pattern Recognition. pp. 652--660 (2017)

\bibitem{qi2017pointnet++}
Qi, C.R., Yi, L., Su, H., Guibas, L.J.: Pointnet++: Deep hierarchical feature learning on point sets in a metric space. Advances in Neural Information Processing Systems  \textbf{30} (2017)

\bibitem{radford2021learning}
Radford, A., Kim, J.W., Hallacy, C., Ramesh, A., Goh, G., Agarwal, S., Sastry, G., Askell, A., Mishkin, P., Clark, J., et~al.: Learning transferable visual models from natural language supervision. In: International Conference on Machine Learning. pp. 8748--8763. PMLR (2021)

\bibitem{schuster1997bidirectional}
Schuster, M., Paliwal, K.K.: Bidirectional recurrent neural networks. IEEE Transactions on Signal Processing  \textbf{45}(11),  2673--2681 (1997)

\bibitem{sepas2022deep}
Sepas-Moghaddam, A., Etemad, A.: Deep gait recognition: A survey. IEEE Transactions on Pattern Analysis and Machine Intelligence  \textbf{45}(1),  264--284 (2022)

\bibitem{shen2023lidargait}
Shen, C., Fan, C., Wu, W., Wang, R., Huang, G.Q., Yu, S.: Lidargait: Benchmarking {3D} gait recognition with point clouds. In: Proceedings of the IEEE/CVF Conference on Computer Vision and Pattern Recognition. pp. 1054--1063 (2023)

\bibitem{takemura2018multi}
Takemura, N., Makihara, Y., Muramatsu, D., Echigo, T., Yagi, Y.: Multi-view large population gait dataset and its performance evaluation for cross-view gait recognition. IPSJ Transactions on Computer Vision and Applications  \textbf{10},  1--14 (2018)

\bibitem{teepe2021gaitgraph}
Teepe, T., Khan, A., Gilg, J., Herzog, F., H{\"o}rmann, S., Rigoll, G.: Gaitgraph: Graph convolutional network for skeleton-based gait recognition. In: 2021 IEEE International Conference on Image Processing (ICIP). pp. 2314--2318. IEEE (2021)

\bibitem{vaswani2017attention}
Vaswani, A., Shazeer, N., Parmar, N., Uszkoreit, J., Jones, L., Gomez, A.N., Kaiser, {\L}., Polosukhin, I.: Attention is all you need. Advances in Neural Information Processing Systems  \textbf{30} (2017)

\bibitem{wang2010chrono}
Wang, C., Zhang, J., Pu, J., Yuan, X., Wang, L.: Chrono-gait image: A novel temporal template for gait recognition. In: Computer Vision--ECCV 2010: 11th European Conference on Computer Vision, Heraklion, Crete, Greece, September 5-11, 2010, Proceedings, Part I 11. pp. 257--270. Springer (2010)

\bibitem{wang2019pseudo}
Wang, Y., Chao, W.L., Garg, D., Hariharan, B., Campbell, M., Weinberger, K.Q.: Pseudo-lidar from visual depth estimation: Bridging the gap in {3D} object detection for autonomous driving. In: Proceedings of the IEEE/CVF Conference on Computer Vision and Pattern Recognition. pp. 8445--8453 (2019)

\bibitem{wu2017rgb}
Wu, A., Zheng, W.S., Yu, H.X., Gong, S., Lai, J.: Rgb-infrared cross-modality person re-identification. In: Proceedings of the IEEE International Conference on Computer Vision. pp. 5380--5389 (2017)

\bibitem{yang2022vision}
Yang, J., Duan, J., Tran, S., Xu, Y., Chanda, S., Chen, L., Zeng, B., Chilimbi, T., Huang, J.: Vision-language pre-training with triple contrastive learning. In: Proceedings of the IEEE/CVF Conference on Computer Vision and Pattern Recognition. pp. 15671--15680 (2022)

\bibitem{yang2024depth}
Yang, L., Kang, B., Huang, Z., Xu, X., Feng, J., Zhao, H.: Depth anything: Unleashing the power of large-scale unlabeled data. arXiv preprint arXiv:2401.10891  (2024)

\bibitem{ye2019modality}
Ye, M., Lan, X., Leng, Q.: Modality-aware collaborative learning for visible thermal person re-identification. In: Proceedings of the 27th ACM International Conference on Multimedia. pp. 347--355 (2019)

\bibitem{ye2020cross}
Ye, M., Lan, X., Leng, Q., Shen, J.: Cross-modality person re-identification via modality-aware collaborative ensemble learning. IEEE Transactions on Image Processing  \textbf{29},  9387--9399 (2020)

\bibitem{ye2020dynamic}
Ye, M., Shen, J., J.~Crandall, D., Shao, L., Luo, J.: Dynamic dual-attentive aggregation learning for visible-infrared person re-identification. In: Computer Vision--ECCV 2020: 16th European Conference, Glasgow, UK, August 23--28, 2020, Proceedings, Part XVII 16. pp. 229--247. Springer (2020)

\bibitem{yu2006framework}
Yu, S., Tan, D., Tan, T.: A framework for evaluating the effect of view angle, clothing and carrying condition on gait recognition. In: 18th International Conference on Pattern Recognition (ICPR'06). vol.~4, pp. 441--444. IEEE (2006)

\bibitem{zhao2021point}
Zhao, H., Jiang, L., Jia, J., Torr, P.H., Koltun, V.: Point transformer. In: Proceedings of the IEEE/CVF International Conference on Computer Vision. pp. 16259--16268 (2021)

\end{thebibliography}

\end{document}